\documentclass{article}

\usepackage[main, final]{neurips_2026}

\usepackage[utf8]{inputenc} 
\usepackage[T1]{fontenc}    
\usepackage{hyperref}       
\usepackage{url}            
\usepackage{booktabs}       
\usepackage{amsfonts}       
\usepackage{nicefrac}       
\usepackage{microtype}      
\usepackage{xcolor}         

\usepackage{amsmath} 
\usepackage{bm}  
\usepackage{makecell} 
\usepackage{multirow} 
\usepackage{caption} 

\usepackage{graphicx}
\usepackage{enumitem}

\usepackage{verbatim}

\newcommand{\etc}{\textit{etc}.}

\newcommand{\nickname}{FoundSurface}

\title{\nickname{}: Feedforward Ray-Based 3D Scene Surface Reconstruction from Unposed Images}

\author{%
Shenxing Wei\textsuperscript{1}\quad Jinxi Li\textsuperscript{1}\quad Zihui Zhang\textsuperscript{1}\quad Ajay Kumar\textsuperscript{2}\quad Wei Wang \textsuperscript{3} \thanks{Corresponding authors.} \quad
  Bo Yang\textsuperscript{1} \footnotemark[1] \\[0.5em]
  \textsuperscript{1}vLAR Group, \textsuperscript{2}DSAI, The Hong Kong Polytechnic University;
  \textsuperscript{3}Regressor, Inc.\\
  \texttt{\{shenxing.wei,jinxi.li,zihui.zhang\}@connect.polyu.hk, bo.yang@polyu.edu.hk} \\
}

\begin{document}

\maketitle
\begin{center}
    \vspace{-25pt}
    \centering
    \includegraphics[width=0.9\linewidth]{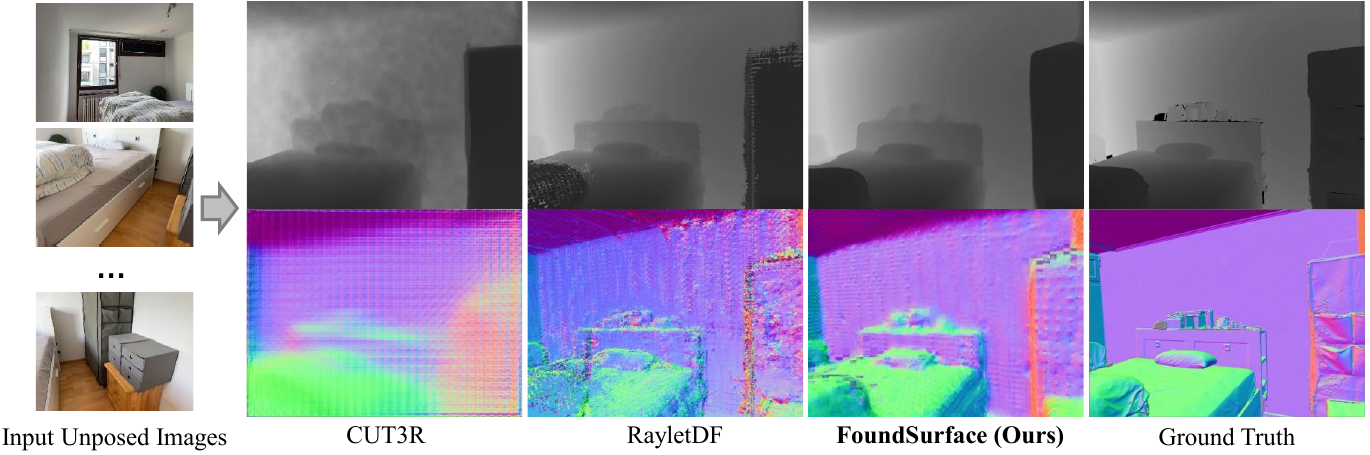}
    \captionof{figure}{We introduce a large feed-forward model \nickname{} to recover continuous 3D scene surfaces from an arbitrary number of images without needing poses.}
    \label{fig:opening}
    \vspace{5pt}
\end{center}

\begin{abstract}
Accurate 3D surface reconstruction from unposed images remains a fundamental challenge in computer vision. Recent feed-forward pointmap methods eliminate the need for camera poses but inherently produce discrete, sparse representations. In this paper, we present a generalizable pipeline that directly recovers continuous, high-fidelity 3D surfaces from unposed images by seamlessly integrating pointmap- and ray-based representations. Our approach first extracts explicit geometric priors using a pre-trained pointmap model. A query-view-centric selection module then identifies geometrically relevant reference views via a robust surfel voting mechanism. Next, a cross-attention regressor bridges target query rays with reference features to estimate a continuous coarse surface. Lastly, a raylet-based module aggregates local 3D features to carve out high-frequency details and analytical normals. Trained as a single foundation model on 9 diverse datasets, our method demonstrates exceptional zero-shot generalization on 6 unseen datasets, clearly surpassing state-of-the-art baselines in geometric accuracy and detail preservation. Code and data are available at: \textcolor{blue}{\url{https://vlar-group.github.io/FoundSurface.html}}
\end{abstract}

\section{Introduction}\label{sec:intro}

Accurate 3D surface reconstruction from multiple images plays a key role in emerging applications such as embodied AI and augmented reality. In recent years, significant progress has been made in the field of 3D reconstruction through the development of various 3D representation techniques \cite{Wang2024}. These include coordinate-based methods such as radiance fields (NeRF) \cite{Mildenhall2020}, and point-based methods like 3D Gaussian Splatting (3DGS) \cite{Kerbl2023}, and their variants \cite{Gao2022,Chen2024c}. While these approaches achieve impressive results in novel view synthesis, they usually rely on highly accurate camera poses, which are often unavailable in practice, and typically struggle to recover precise 3D surfaces. Furthermore, they generally require time-consuming per-scene optimization rather than functioning as generalizable, feed-forward models \cite{Zhang2025}.

Recently, the pioneering feedforward method DUSt3R \cite{Wang2024b} has enabled precise 3D geometry reconstruction by directly predicting pixel-aligned pointmaps from unposed image pairs. Subsequent methods \cite{Zhang2025c} have extended this pointmap representation to handle multiple views \cite{Wang2025a,Yang2025b}, to improve computational efficiency \cite{Shen2026,Liu2025}, \etc{}. Leveraging powerful Transformer architectures and large-scale training datasets, these approaches achieve excellent accuracy in recovering surface points from image collections. However, the reconstructed pointmaps are inherently discrete and sparse, lacking the continuous and fine-grained geometric details required for many downstream applications, such as robot manipulation and scene editing.

Another recent line of work focuses on ray-based representations such as RayDF \cite{Liu2023a} and RayletDF \cite{Wei2025}. Their key innovation is learning a ray-surface distance field for any query light ray conditioned on visual inputs, thus enabling the recovery of continuous and high-fidelity surfaces. However, they still rely on accurate camera poses or sparse point clouds as inputs to reconstruct detailed surfaces.

\begin{figure}[t]
\centering
  \includegraphics[width=.99\linewidth]{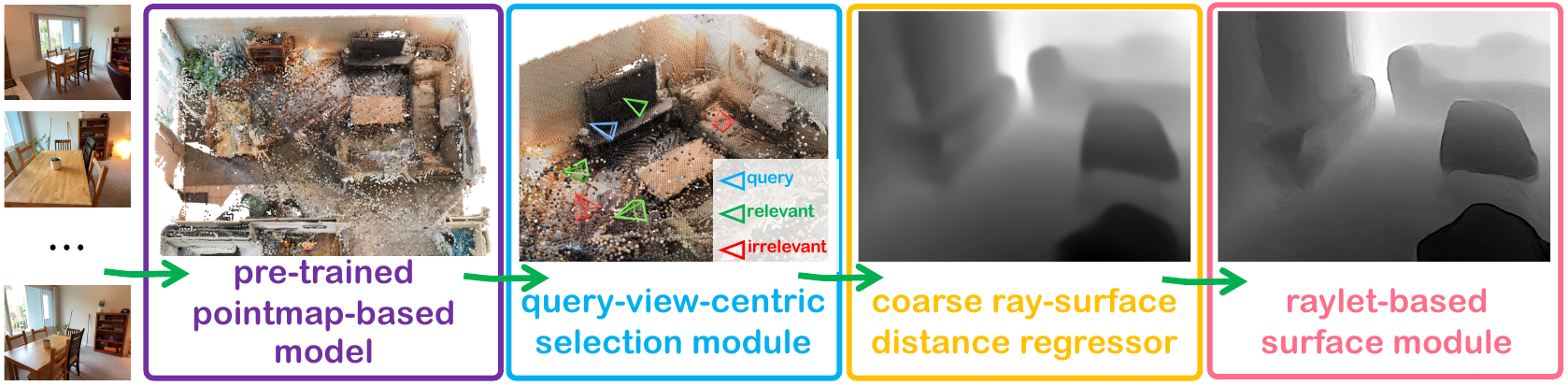}
\caption{The general workflow and components of our framework.}
\label{fig:overview}
\vspace{-0.6cm}
\end{figure}

In this paper, we present a generalizable, feed-forward 3D surface reconstruction pipeline that takes only a set of unposed images as input and directly recovers continuous, high-fidelity 3D surfaces. We achieve this by seamlessly integrating pointmap- and ray-based representations. In particular, as illustrated in Figure \ref{fig:overview}, our pipeline comprises four key components: 1) a pre-trained pointmap-based model such as VGGT \cite{Wang2025a}, which extracts visual tokens and generates pixel-aligned pointmaps along with estimated camera poses; 2) a query-view-centric selection module designed to identify geometrically relevant reference views; 3) a coarse ray-surface distance regressor utilizing cross-attention layers; and 4) a raylet-based surface module for refining surface details and normals.

Among the four modules of our pipeline, the core is the second and third modules, which seamlessly integrate pointmap- and ray-based representations, effectively estimating continuous surfaces for any query views conditioned on the recovered pointmaps from unposed input images. In contrast, directly applying RayletDF to VGGT pointmaps can lead to inaccurate raylet initialization, resulting in multi-surface artifacts and missing regions.

Our pipeline is based on large Transformer and standard neural layers with sufficient capacity, without relying on domain specific assumptions. Therefore, it is rather suitable to consume various indoor and outdoor datasets. We train a single \textbf{found}ation 3D \textbf{surface} model on a mixture of 9 public datasets with depth annotations, including BlendMVS \cite{Yao2020}, WildRGB \cite{Xia2024}, ScanNet \cite{Dai2017}, HyperSim \cite{Roberts2021}, MVS-Synth \cite{Huang2018}, ARKitScenes \cite{Baruch2021}, S3DIS \cite{Armeni2017}, SceneNN \cite{Hua2016}, and GTASfM \cite{Wang2020}. Our model is termed \textbf{\nickname{}} and Figure \ref{fig:opening} shows qualitative results. Our contributions are: 

\begin{itemize}[leftmargin=*]
\setlength{\itemsep}{1pt}
\setlength{\parsep}{1pt}
\setlength{\parskip}{1pt}
  \vspace{-0.2cm}
    \item 
    We introduce a large feed-forward model to recover continuous 3D scene surfaces from an arbitrary number of images without needing poses.
    \item We design a query-view-centric selection module together with a cross attention based fusion module to seamlessly integrate pointmap- and ray-based 3D representations.    
    \item We train a single foundation model and demonstrate superior 3D surface reconstruction accuracy across multiple datasets, clearly surpassing existing baselines.  
\end{itemize}

\section{Related Work}\label{sec:liter}

To recover 3D surfaces from images, classical SfM \cite{Ozyesil2017} and SLAM \cite{Cadena2016} systems extract sparse point clouds that can be smoothed into continuous surfaces, though they often fail to capture fine details. Early deep learning methods utilized explicit 3D representations such as voxel grids \cite{Chan2016,Yang2018,Yang2020}, point clouds \cite{Fan2017}, octrees \cite{Tatarchenko2017}, and meshes \cite{Kato2017}. Nevertheless, the quality of these discrete structures is inherently bottlenecked by memory footprints and resolution constraints. Comprehensive surveys of these methods are available in \cite{Berger2017a,Han2019}.

\textbf{Implicit Representations for Surface Reconstruction}:
To bypass the resolution limits of explicit 3D formats, implicit representations use simple MLPs to parameterize continuous 3D shapes \cite{Fried2020}. Techniques such as occupancy fields (OF) \cite{Chen2019g,Mescheder2019}, un/signed distance fields (U/SDF) \cite{Chibane2020a,Park2019,Wang2022,Deng2026}, vector fields (VF) \cite{Yang2023,Rella2024}, and NeRF \cite{Mildenhall2020,Trevithick2021} operate by mapping 3D coordinates to local geometric or visual properties. While these approaches and their variants \cite{Wang2024} yield highly accurate surface reconstructions and 2D renderings, they rely on dense 3D spatial sampling and exhaustive network evaluations per view and often require images with accurate poses as input. In this paper, our \nickname{} learns a feedforward model from unposed images, directly regressing surface points for any query viewing rays in single pass. 

\textbf{Point-based Representations for Surface Reconstruction}: A recent breakthrough is 3D Gaussian Splatting (3DGS) \cite{Kerbl2023}, which models environments as a set of 3D Gaussians with optimizable positions, covariances, and colors. Supported by a custom tile-based rasterizer, 3DGS delivers real-time 2D view synthesis. Nevertheless, because it inherently lacks tight geometric constraints, 3DGS falls short in extracting fine surface details and generating highly accurate depth maps. While a multitude of recent works \cite{Chen2024,Fan2024,Wu2024c,Guedon2024,Huang2024,Chung2024,Xu2025,Li2024,Huang2025,Yu2024a,Xiang2024,Ouyang2024,Yu2024,Song2024,Chen2024a,Zhang2024,Gao2025} have attempted to enforce surface alignment and improve depth quality, they generally require time-consuming per-scene optimization and images with accurate poses as input.

Another breakthrough is DUSt3R \cite{Wang2024b} which demonstrates 3D geometry can be represented by pixel-aligned pointmaps and recovered from unposed image pairs in a single forward pass. A series of succeeding methods \cite{Zhang2025c,Leroy2024,Cabon2025,Elflein2025,Chen2025,Wang2025a,Wang2025,Yang2025b,Jang2025,Zhang2025d,Li2026,Xie2026,Jin2026,Zhuo2026,Lan2026,Fang2026,Li2026b} extend it from different perspectives, including handling more views , modeling videos, processing faster, \etc{}. Although these methods achieve impressive results, they are limited to recovering discrete and sparse point clouds, falling short of producing continuous, high-fidelity surfaces. 

\textbf{Ray-based Representations for Surface Reconstruction}: Another line of work models 3D shapes through ray-based neural functions. By taking individual rays as input, these models directly regress either RGB values \cite{Liu2021a,Feng2021,Ost2022,Suhail2022,Cao2023} or explicit surface points \cite{Feng2022,Aumentado-Armstrong2022,Liu2023a,Wei2025,Kulkarni2022,Sundt2023,Jin2024,Sundt2024,Yin2024}. Despite their impressive rendering efficiency and surface accuracy, most of these methods require posed images as input and struggle to generalize across diverse 3D scenes. In this paper, we leverage the robust surface representation capabilities of RayDF \cite{Liu2023a} and RayletDF \cite{Wei2025} and seamlessly integrate them with pointmap-based representations. This synergistic approach enables feedforward 3D surface reconstruction directly from sets of images.

\section{\nickname{}}\label{sec:method}

\nickname{} recovers continuous high-fidelity 3D scene surfaces through a coarse-to-fine, discrete-to-continuous progression. Given a set of unposed and uncalibrated multi-view images $\mathcal{I}=\{\mathbf{I}_i\in\mathbb{R}^{H\times W\times 3}\ | \ i=1,...,K \}$, we first extract an explicit geometric prior using a pre-trained pointmap-based model elaborated in Section \ref{sec:pointmap_model}. Next, we filter out redundant information by selecting geometrically relevant views for any target query in Section \ref{sec:selection_module}. We then transition from 3D points to rays by regressing an initial, continuous coarse surface, which resolves global geometry and occlusions detailed in Section \ref{sec:coarse_regressor}. Finally, we refine this coarse surface using a raylet-based module that aggregates high-resolution local 3D features to recover fine-grained details, yielding the final high-fidelity surfaces, discussed in Section \ref{sec:raylet_module}.

\subsection{Pre-trained Pointmap-based Model}
\label{sec:pointmap_model}

This model aims to establish a reliable spatial anchor for building continuous surfaces without camera poses. Given a set of input images $\mathcal{I}=\{\mathbf{I}_i \in \mathbb{R}^{H \times W \times 3}\}$, we employ the pre-trained VGGT model~\cite{Wang2025a} to predict explicit geometric priors. Specifically, the network's prediction heads output a pixel-aligned pointmap $\mathbf{P}_i \in \mathbb{R}^{H \times W \times 3}$, a camera pose matrix $\mathbf{T}_i \in \mathbb{R}^{4 \times 4}$, and an intrinsic matrix $\mathbf{C}_i \in \mathbb{R}^{3 \times 3}$ for each input image $\mathbf{I}_i$. 

During this forward pass, we concurrently retain intermediate patch features to capture rich contextual information. Let $\mathbf{F}_{i, l} \in \mathbb{R}^{h \times w \times d_l}$ denote the patch feature extracted from the $l$-th layer of VGGT, where $h=H/s$ and $w=W/s$ are the patch-level spatial dimensions downsampled from $H$ and $W$ by ratio $s$, and $d_l$ is the channel depth. We specifically collect features from four layers, defined as the set $L = \{4, 11, 17, 23\}$. To prepare compact and geometry-aware visual tokens for subsequent modules, we fuse these multi-scale patch features using a two-layer MLP, denoted as $f_{\mathrm{fuse}}$. The entire extraction and fusion process can be succinctly formulated as:
\begin{equation}
    \{ \mathbf{P}_i, \mathbf{T}_i, \mathbf{C}_i, \{ \mathbf{F}_{i, l} \}_{l \in L} \} \leftarrow \text{VGGT}(\mathbf{I}_i), \quad
    \mathbf{F}_i = f_{\mathrm{fuse}}([\mathbf{F}_{i, 4}, \mathbf{F}_{i, 11}, \mathbf{F}_{i, 17}, \mathbf{F}_{i, 23}])
\end{equation}
where the operator $[\cdot]$ denotes concatenation along the channel dimension, yielding the final fused visual feature $\mathbf{F}_i \in \mathbb{R}^{h \times w \times d_{out}}$. The discrete geometric predictions $\mathbf{P}_i / \mathbf{T}_i / \mathbf{C}_i$ and the fused visual tokens $\mathbf{F}_i$ constitute the foundation for our continuous ray-based formulation.

\begin{figure}[t]
\centering
  \includegraphics[width=.99\linewidth]{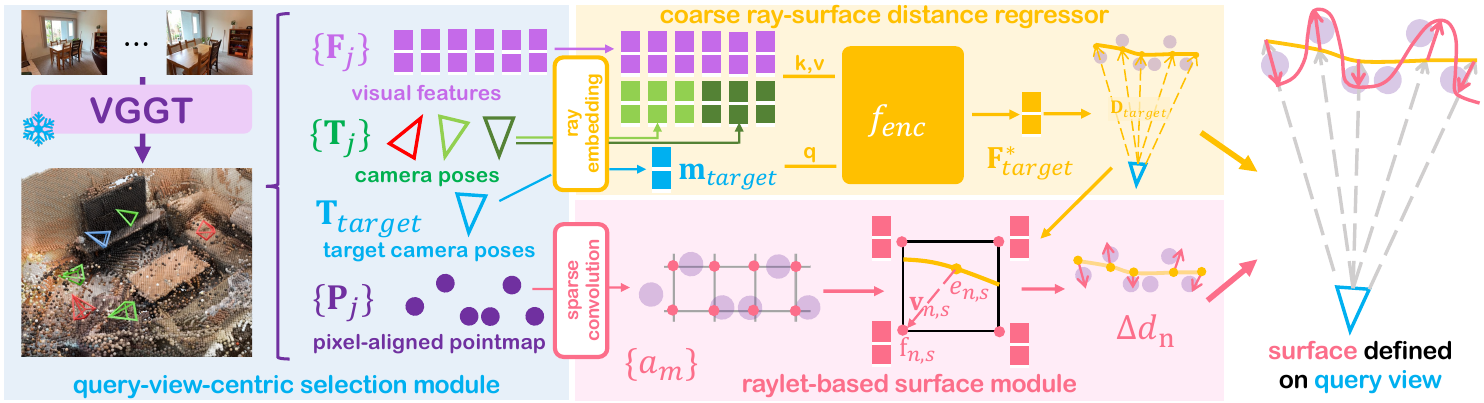}
\caption{Left: The query-view-centric selection module. Middle: The coarse ray-surface distance regressor, which seamlessly integrates pointmap- and ray-based representations. Right: The raylet-based surface module, which further refines the initially estimated surfaces.}
\label{fig:method}
\vspace{-0.4cm}
\end{figure}

\subsection{Query-View-Centric Selection Module}
\label{sec:selection_module}
As shown in the leftmost block of Figure \ref{fig:method}, when reconstructing the geometry of a novel query view, which is defined explicitly by a given target camera pose $\mathbf{T}_{target}$ and $\mathbf{C}_{target}$, intuitively, not all input views are helpful. Views that do not observe the target surfaces will only introduce noise and computational overhead. Therefore, instead of globally querying features of all input views, we design a selection module to identify only the overlapping, geometrically relevant reference views by a \textit{surfel-based} voting mechanism, ensuring the selected views provide reliable geometric coverage.

Specifically, we first establish a canonical geometric reference for this novel query view. We project all available input pointmaps onto the camera plane defined by the target query pose, and apply a standard depth buffer (z-buffer) to render a dense target pseudo-depth map, $\mathbf{D}^{target}$. We then uniformly divide $\mathbf{D}^{target}$ into non-overlapping patches, conceptualized as target surfels.

To evaluate a candidate view $j$ out of all input views, we similarly divide its pointmap $\mathbf{P}_j$ into candidate surfels and project them onto the target view. For each pixel $(x,y)$ within a projected candidate surfel, we compare its depth value $\mathbf{D}^j_{x,y}$ against the canonical target depth $\mathbf{D}^{target}_{x,y}$. The pixel casts a valid match vote according to an indicator function:
\begin{equation}
    v_{x,y} = \mathbb{I} \left( \left| \mathbf{D}^{target}_{x,y} - \mathbf{D}^{j}_{x,y} \right| < \delta_{depth} \right)
\end{equation}
A candidate surfel is considered successfully ``matched'' if the proportion of positive votes $v_{x,y}$ within it exceeds a threshold $\tau_{vote}$. By aggregating these surfel-level results, we compute the percentage of matched surfels for the entire candidate view $j$, yielding a robust, coverage-aware overlap score. Finally, we rank all input views based on this surfel overlap score and retain only the \textit{top-k} candidates. This geometric filtering eliminates irrelevant regions and provides our subsequent attention modules with a highly curated and effective query candidate set: $\{\mathbf{T}_j\}^K_{j=1}, \{\mathbf{C}_j\}^K_{j=1}, \{\mathbf{P}_j\}^K_{j=1}, \{\mathbf{F}_j\}^K_{j=1}$. More implementation details are provided in Appendix~\ref{app:detail_ref_selection}.

\subsection{Coarse Ray-Surface Distance Regressor}
\label{sec:coarse_regressor}

With the relevant views selected, the next challenge is locating the actual surface intersection for the target camera rays. This module aims to seamlessly integrate pointmap- and ray-based representations to directly estimate an initial, continuous coarse surface. To robustly regress this initial surface, we establish a local coordinate system centered at the target query view. As shown in the yellow block of Figure \ref{fig:method}, given the target camera pose $\mathbf{T}_{target}$ and intrinsics $\mathbf{C}_{target}$, we align the candidate views to this target viewpoint and tokenize their pixel-aligned Plücker rays via a convolution layer ($\text{Conv}$) to yield ray embeddings $\mathbf{m}_j$ for each relevant view $j$ as:
\begin{equation}
    \mathbf{m}_j = \text{Conv}\big(\mathcal{R}(\mathbf{T}_j^{rel}, \mathbf{C}_j)\big)\in\mathbb{R}^{hw\times d}, \quad \text{where } \mathbf{T}_j^{rel} = \mathbf{T}_{target}^{-1} \mathbf{T}_j
\end{equation}
Here, $\mathbf{T}_j^{rel}$ defines the relative transformation to the target viewpoint, and $\mathcal{R}(\cdot)$ denotes the operation that casts camera parameters into dense Plücker ray maps. To inject spatial awareness into the visual features $\mathbf{F}_j$ (obtained from the backbone in Section \ref{sec:pointmap_model}), we condition them with the corresponding Plücker ray tokens using an MLP-based network $f_{\mathrm{cond}}$ and get the reference features $\hat{\mathbf{F}}_j$ as:
\begin{equation}
    \hat{\mathbf{F}}_j = f_{\mathrm{cond}}([\mathbf{m}_j, \mathbf{F}_j]),
\end{equation}
where $[\cdot]$ denotes concatenation along the channel dimension. 

Similarly, for the target query view, we cast Plücker rays using its parameters $(\mathbf{I}_{4\times4}, \mathbf{C}_{target})$ and tokenize them into target ray embeddings $\mathbf{m}_{target}$. To aggregate spatially aligned features from the candidates, we employ a Transformer network $f_{\mathrm{enc}}$ consisting of alternating self- and cross-attention layers. Specifically, in the cross-attention mechanism, the target query ray embeddings act as the queries ($\mathbf{Q}$), while the conditioned candidate features $\{\hat{\mathbf{F}}_j\}_{j=1}^K$ act as the keys ($\mathbf{K}$) and values ($\mathbf{V}$). This allows the target query rays to adaptively fetch relevant geometric information:
\begin{equation}
    \mathbf{F}^*_{target} = f_{\mathrm{enc}} \left( \mathbf{Q} = \mathbf{m}_{target}; \ \mathbf{K,V} = \{\hat{\mathbf{F}}_j\}_{j=1}^K \right)
\end{equation}

Decoding the aggregated feature map $\mathbf{F}^*_{target}$ yields a ray distance map $\mathbf{D}^{target}$. This distance map effectively establishes our continuous, albeit coarse, ray-surface representation. Specifically, to locate the 3D surface point $\mathbf{p}_{x,y}$ for a given pixel $(x,y)$ in the target query view, we cast a ray using the known target query camera parameters:
\begin{equation}
    \mathbf{p}_{x,y} = \mathbf{p}_{target} + \mathbf{D}^{target}_{x,y} \cdot \mathbf{r}_{x,y}
\end{equation}
where $\mathbf{p}_{target} \in \mathbb{R}^3$ is the optical center and $\mathbf{r}_{x,y} \in \mathbb{R}^3$ is the normalized ray direction. Both are uniquely determined by the target query pose $\mathbf{T}_{target}$ and intrinsics $\mathbf{C}_{target}$. More implementation details can be found in Appendix~\ref{app:detail_coarse_regressor}. 

While this coarse surface successfully captures the overall scene geometry and resolves occlusions, it inherently lacks fine-grained, high-frequency local details due to the patch-level resolution of the visual tokens. Therefore, we treat these coarse surface points and their corresponding ray directions as a set of sampled raylets conceptualized by RayletDF \cite{Wei2025}, denoted as $\{\mathbf{l}_n = (\mathbf{p}_n, \mathbf{r}_n) \}_{n=1}^{N}$, which serve as the geometric foundation for our subsequent raylet-based surface module to refine the details.

\subsection{Raylet-based Surface Module}
\label{sec:raylet_module}
To elevate our coarse surface to a high-fidelity continuous representation, we must compensate for the missing local geometric details. We achieve this by refining the coarse raylets $\{\mathbf{l}_n\}_{n=1}^N$ through the aggregation of high-resolution local 3D features. 

We first voxelize the initial pointmaps $\{\mathbf{P}_j\}^K_{j=1}$ and index the corresponding visual features $\{\mathbf{F}_j\}^K_{j=1}$ obtained in Section \ref{sec:coarse_regressor}. We then utilize a 3D sparse convolution network (SPConv)~\cite{spconv2022} to process these indexed features, constructing a robust 3D local feature volume. 
As shown in the rightmost block of Figure 3, the sparse convolution network produces a set of spatial anchors and their associated local features, denoted as $\mathcal{A} = \{\mathbf{a}_m\}_{m=1}^M$. Here, the anchor points are generated through sparse convolution over the voxelized 3D representation, rather than being directly sampled from the original VGGT pointmaps. Each anchor is associated with a learned local geometric feature, which is subsequently used to refine nearby raylets.

For each coarse raylet $\mathbf{l}_n = (\mathbf{p}_n, \mathbf{r}_n)$ generated from the previous module, we search for its $S$-nearest anchor points in 3D space, yielding a local neighborhood $\mathcal{A}_n = \{(\mathbf{a}_{n,s}, \mathbf{f}_{n,s})\}_{s=1}^S$. To make the network highly sensitive to local geometry, we explicitly calculate the spatial relationships between the coarse raylet and its local anchors. Specifically, we extract the Euclidean distances $e_{n,s} = \|\mathbf{p}_n - \mathbf{a}_{n,s}\|_2$ and the relative direction vectors $\mathbf{v}_{n,s} = (\mathbf{p}_n - \mathbf{a}_{n,s}) / e_{n,s}$. 

Finally, a refinement network processes the local anchor features alongside these explicit geometric relationships to regress a precise distance residual $\Delta d_n$. This aggregation is formulated as:
\begin{equation}
    \Delta d_n = f_{\mathrm{refine}} \Bigg( \bigg[ \mathop{\text{MEAN}}\limits_{s=1}^S (\mathbf{f}_{n,s}), \, e_{n,1}, \mathbf{v}_{n,1}, \dots, e_{n,S}, \mathbf{v}_{n,S} \bigg] \Bigg)
\end{equation}
where $[\cdot]$ denotes feature concatenation, and $\text{MEAN}$ represents the mean-pooling function applied exclusively to the visual features of the $K$ local neighbors. The final refined surface point is obtained by updating the coarse raylet:
\begin{equation}
    \mathbf{p}^*_n = \mathbf{p}_n + \Delta d_n \cdot \mathbf{r}_n
\end{equation}
This refinement dynamically carves out the high-frequency details, yielding a continuous, high-fidelity 3D surface. As a byproduct of this continuous ray-based formulation, our method natively supports the analytical derivation of surface normals. Given the input raylet and the final refined distance, we can easily compute a closed-form normal vector, effectively circumventing the noise typical of numerical approximations. We defer the full derivation and other details to the Appendix~\ref{app_normal_dev}.

\subsection{Training}
\label{sec:training}
We employ a decoupled two-stage training strategy to train our core modules. 

\textbf{Stage 1: Coarse Ray-Surface Distance Regression.} In this stage, our goal is to establish an accurate global geometric framework. We train the coarse ray-surface distance regressor to predict the continuous coarse depth map $\mathbf{D}^{target}$. Supervision is provided by the aligned ground truth depth $\mathbf{D}_{gt}$, using a loss function that combines a pixel-wise $\ell_1$ distance with a spatial gradient loss to enforce surface consistency:
\begin{equation}
    \ell_{coarse} = |\mathbf{D}_{gt} - \mathbf{D}^{target}| + \|\nabla \mathbf{D}_{gt} - \nabla \mathbf{D}^{target} \|
\end{equation}

\textbf{Stage 2: Raylet-based Surface Refinement.} Once the coarse geometry is reasonably established, we freeze or isolate the first-stage predictions and focus entirely on training the raylet-based surface module. To capture high-frequency geometric details without destabilizing the learned global structure, we predict a residual distance map $\Delta \mathbf{D}$. During this stage, we apply a stop-gradient operation to the coarse predictions, formulating the fine-stage loss as:
\begin{equation}
    \ell_{fine} = \Big| \mathbf{D}_{gt} - \big( \text{sg}(\mathbf{D}^{target}) + \Delta \mathbf{D} \big) \Big|
\end{equation}
where $\text{sg}(\cdot)$ denotes the stop-gradient operator. Here, the global residual map $\Delta \mathbf{D}$ is formed by aggregating the local raylet residuals corresponding to each pixel, i.e., $\Delta \mathbf{D} = \{ \Delta d_n \}_{n=1}^N$, where $\Delta d_n$ is computed following the local feature aggregation mechanism detailed in Section \ref{sec:raylet_module}. This two-stage training paradigm effectively decouples global topology localization from local surface refinement, thereby enabling high-fidelity 3D reconstructions. More details are in Appendix~\ref{app_train}.

\begin{table}[t]\tabcolsep=0.1cm 
\centering
\captionsetup{skip=2pt}
\caption{Quantitative results of estimated ray-surface distances.}
\resizebox{1\textwidth}{!}{
\begin{tabular}{@{}r|ccccc|ccccc|ccccc@{}}
    \toprule
   &\multicolumn{5}{c|}{\textit{test on} $\rightarrow$ ScanNet++}&\multicolumn{5}{c|}{\makecell{\textit{test on} $\rightarrow$ 7Scenes}}&\multicolumn{5}{c}{\makecell{\textit{test on} $\rightarrow$ MultiScan}}\\
   \cmidrule{2-16}
   &ADE$\downarrow$ &RMSE$\downarrow$ &Abs-Rel$\downarrow$&Sq-Rel$\downarrow$&$\delta\uparrow $&ADE$\downarrow$&RMSE$\downarrow$&Abs-Rel$\downarrow$&Sq-Rel$\downarrow$&$\delta\uparrow$&
   ADE$\downarrow$&RMSE$\downarrow$&Abs-Rel$\downarrow$&Sq-Rel$\downarrow$&$\delta\uparrow$ \\
   \midrule
   Projection~\cite{Wang2025a}&
   0.580&0.959&0.360&0.554&0.636& 0.188&0.391&0.091&0.078&0.887&0.566&0.767&0.462&0.521&0.510 \\
   PriorDA~\cite{wang2025depthprior}&
   0.198&0.290&0.123&0.101&0.868 &0.154&0.266&0.074&0.034&0.913    &0.384&0.471&0.307&0.310&0.646\\
    CUT3R~\cite{Wang2025}&
   0.335&0.446&0.222&0.173&0.703 &0.197&0.305&0.105&0.054&0.878    &0.452&0.530&0.454&0.410&0.463\\
   FLARE~\cite{Zhang2025d}&
   0.312&0.495&0.157&0.156&0.771 &0.228 &0.363 &0.112 &0.066 &0.841    &0.396 &0.589 &0.305 &0.343 &0.677 \\
    RayletDF~\cite{Wei2025}&
   0.179&0.334&0.119&0.205&0.896 &0.158&0.295&0.084&0.049&0.907    &0.331&0.447&0.306&0.425&0.719\\
    Ours \textit{w/o refine}&
   0.162&0.270&0.097&0.063&0.894 &0.130&\textbf{0.230}&0.068&\textbf{0.029}&\textbf{0.945}    &0.233&0.309&0.222&0.147&0.737\\
   \textbf{Ours} &
   \textbf{0.132}&\textbf{0.259}&\textbf{0.079}&\textbf{0.067}&\textbf{0.910}& \textbf{0.123}&0.233&\textbf{0.064}&0.031&0.943 &\textbf{0.205}&\textbf{0.287}&\textbf{0.190}&\textbf{0.137}&\textbf{0.774}\\
   \midrule
    &\multicolumn{5}{c|}{\textit{test on} $\rightarrow$ Replica}&\multicolumn{5}{c|}{\makecell{\textit{test on} $\rightarrow$ MuSHRoom}}&\multicolumn{5}{c}{\makecell{\textit{test on} $\rightarrow$ Tanks\&Temples}}\\
   \cmidrule{2-16}
    Projection~\cite{Wang2025a}&
    0.495&1.093&0.201&0.453&0.802 &1.192&1.596&0.759&1.539&0.333 &3.135&6.044&0.289&5.825&0.771\\
   PriorDA~\cite{wang2025depthprior}&
   0.116&0.189&0.045&0.018&0.974 &0.341&0.506&0.309&0.476&0.760 &1.572&\textbf{2.586}&\textbf{0.136}&\textbf{0.970}&0.850\\
    CUT3R~\cite{Wang2025}&
   0.365&0.473&0.149&0.106&0.772 &0.689&0.868&0.610&1.028&0.485    &4.982&6.482&0.579&8.779&0.444\\
    FLARE~\cite{Zhang2025d}
   &0.307 &0.453 &0.115 &0.063 &0.843   &0.553 &0.882 &0.323 &0.528 &0.649  &2.014 &3.390 &0.143 &1.218& 0.803\\
    RayletDF~\cite{Wei2025}&
   0.071&0.179&0.032&0.023&0.975 &0.374&0.640&0.390&0.996&0.795    &3.970&7.940&0.442&10.93&0.748\\
    Ours \textit{w/o refine}&
   0.137&0.215&0.053&0.021&0.976 &0.350&0.511&0.299&0.430&0.734  &1.622&3.144&0.183&1.929&0.865\\
   \textbf{Ours}  & 
   \textbf{0.064}&\textbf{0.165}&\textbf{0.027}&\textbf{0.015}&\textbf{0.981} &\textbf{0.254}&\textbf{0.458}&\textbf{0.232}&\textbf{0.429}&\textbf{0.822 }&\textbf{1.233}&2.977&{0.144}&2.012&\textbf{0.908}\\
\bottomrule
\end{tabular}}
\label{tab:exp_ray_surface_dis_esti_pc} \vspace{-0.5cm}
\end{table}

\section{Experiments}\label{sec:exp}
\textbf{Training Datasets}:
We train our model on a mixture of 9 datasets, including the training splits of ScanNet~\cite{Dai2017}, GTASfM~\cite{Wang2020} and ARKitScenes~\cite{Baruch2021}, together with the full datasets of BlendMVS~\cite{Yao2020}, WildRGB~\cite{Xia2024}, HyperSim~\cite{Roberts2021}, MVS-Synth~\cite{Huang2018}, S3DIS~\cite{Armeni2017} and  SceneNN~\cite{Hua2016} . 
For evaluation, we use ScanNet++~\cite{Yeshwanth2023}, 7Scenes~\cite{shotton2013scene}, MultiScan~\cite{Mao2022}, Mushroom~\cite{ren2024mushroom}, Replica~\cite{straub2019replica}, and Tanks\&Temples~\cite{knapitsch2017tanks}. 
Detailed dataset splits are provided in Appendix~\ref{app_dataset}.

\textbf{Baselines}: 
We compare our approach against five representative baselines: \textbf{1) VGGT Projection}~\cite{Wang2025a}, evaluated via the direct projection of its pointmaps; \textbf{2) PriorDA}~\cite{wang2025depthprior}, a state-of-the-art depth completion model applied to VGGT pointmaps; \textbf{3) CUT3R}~\cite{Wang2025}, a feed-forward model utilizing state tokens for novel query views; \textbf{4) FLARE}~\cite{Zhang2025d}, the latest feed-forward Gaussian splatting model; and \textbf{5) RayletDF}~\cite{Wei2025}, a state-of-the-art generalizable ray-based method.

\textbf{Metrics}: We evaluate surface reconstruction by comparing the predicted ray--surface distance against the ground-truth distance. We report the absolute distance error (\textbf{ADE}, in meters), together with four commonly used metrics~\cite{Turkulainen2025}: root mean squared error (\textbf{RMSE}), absolute relative distance (\textbf{Abs Rel}), squared relative distance (\textbf{Sq Rel}), and the Threshold accuracy ($\delta \leq 1.25$) which measures the percentage of predictions that fall within the specified threshold of the ground truth. 

For normal evaluation, we follow the metrics used in normal estimation~\cite{bae2024rethinking}. We report both the mean and median angular errors on normal. Additionally, we evaluate performance by measuring the percentage of pixels with an angular error below thresholds of 11.25°, 22.5°, and 30.0°. Detailed computation of all metrics is in Appendix \ref{app_metric_ray}\&\ref{app_metric_norm}.

\subsection{Evaluation of Out-of-domain Reconstruction}
\label{sec:outdomain_eval}
In this setting, our model takes 50 unposed input views and predicts ray--surface distances for query rays. For each scene, we uniformly sample 50 images from the test RGB sequence as input views. We further ablate the number of input views in Tables~\ref{tab:exp_ray_surface_dis_sparse}\&~\ref{tab:exp_ray_surface_dis_dense}.

We evaluate on ScanNet++, 7Scenes, MultiScan, MuSHRoom, and Tanks\&Temples. These datasets are excluded from both the VGGT and our training datasets, thereby forming a strictly out-of-domain evaluation setting.
For \textbf{VGGT Projection}~\cite{Wang2025a}, we directly project the pointmaps predicted by VGGT into the query view using Open3D~\cite{Zhou2018}. We further combine VGGT projection with \textbf{PriorDA}~\cite{wang2025depthprior}, a state-of-the-art depth completion model, to fill missing regions after projection. For the feed-forward reconstruction model \textbf{CUT3R}~\cite{Wang2025}, we first forward all input views and then query the scene state token to predict query-view point coordinates. For the feed-forward Gaussian splatting model \textbf{FLARE}~\cite{Zhang2025d}, we render query-view depth maps from the predicted 3D Gaussians. For the ray-based method \textbf{RayletDF}~\cite{Wei2025}, which is originally optimized for Gaussians and point clouds, we retrain it on VGGT-predicted pointmaps from our training datasets for a fair comparison. Standard deviation results are provided in Appendix~\ref{app:std}.

\begin{figure}[t]
  \centering
  \captionsetup{skip=2pt}
  \includegraphics[width=0.9\linewidth]{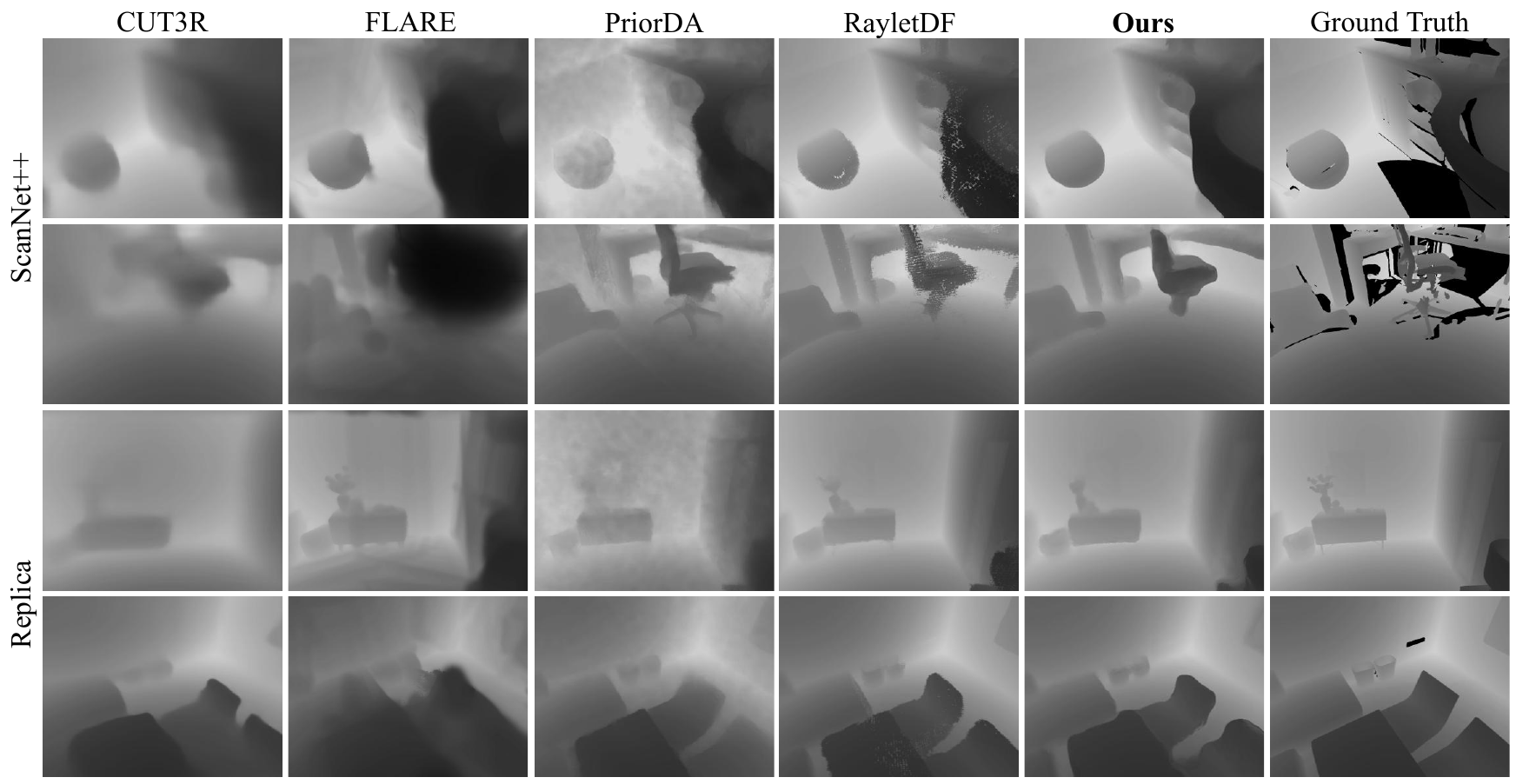}
  \centering
  \caption{Qualitative results of our method and baselines for ray distance on novel views.}
  \vspace{-0.3cm}
  \label{fig:exp_qualitative_raydistance}
\end{figure}

\textbf{Results \& Analysis: }
Table~\ref{tab:exp_ray_surface_dis_esti_pc} and Figure~\ref {fig:exp_qualitative_raydistance} report quantitative and qualitative comparisons on ray-surface distance estimation for all query rays of test images. To assess the contribution of our Raylet-based refinement module, we also report the performance of our coarse ray-surface distance regressor in Table~\ref{tab:exp_ray_surface_dis_esti_pc}, denoted as \textit{w/o refine}.

Our method achieves the best ADE across all out-of-domain datasets, outperforming all baselines by a large margin. For example, on ScanNet++, our ADE is 0.132m compared with 0.198m for PriorDA. These results demonstrate the strong generalization ability of \nickname{} on unseen datasets.  Compared with the state-token-based model CUT3R, our method preserves finer geometric details, as illustrated in Figure~\ref{fig:exp_qualitative_raydistance}. Removing the proposed refinement module substantially degrades reconstruction accuracy, confirming its effectiveness. More results are provided in Appendix \ref{app:add_vis}.

\subsection{Evaluation of Surface Normal Accuracy}
\label{sec:normal_eval}

We further evaluate the normal estimation performance of \nickname{} and all baselines. Since some baselines, such as PriorDA, cannot directly produce analytical surface normals, we additionally compute normals from the predicted depth maps using numerical differentiation for reference.

\begin{figure}[t]
  \centering
  \captionsetup{skip=2pt}
  \includegraphics[width=0.9\linewidth]{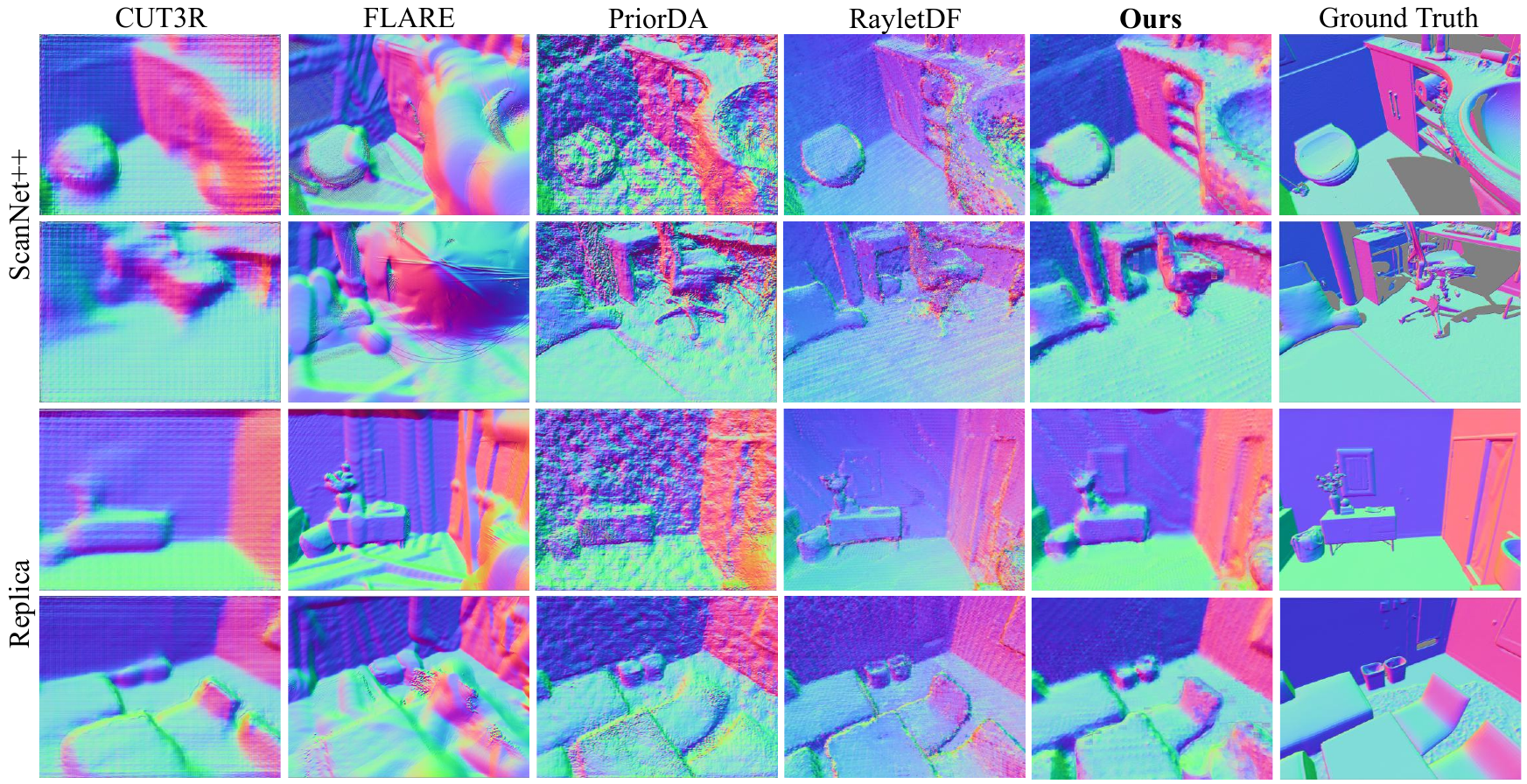}
  \centering
  \caption{Qualitative results of our method and baselines for normal estimation on novel views.}
  \vspace{-0.5cm}
  \label{fig:exp_qualitative_normal}
\end{figure}

\textbf{Results \& Analysis.}
As shown in Table~\ref{tab:exp_ray_surface_normal_esti} and Figure~\ref{fig:exp_qualitative_normal}, our method achieves the best performance using only depth supervision, demonstrating the effectiveness of \nickname{} as a robust surface representation. Additional qualitative comparisons are provided in Appendix~\ref{app:normal_cmp}.

\begin{table}[t]\tabcolsep=0.12cm 
\centering
\captionsetup{skip=2pt}
\caption{Quantitative results of estimated normals. 
}
  \resizebox{0.7\textwidth}{!}{
\begin{tabular}{@{}r|ccccc|ccccc@{}}
    \toprule
   &\multicolumn{5}{c|}{\textit{test on} $\rightarrow$ ScanNet++}&\multicolumn{5}{c}{\makecell{\textit{test on} $\rightarrow$ Replica}}\\
   \cmidrule{2-11}
   &mean$\downarrow$ &med$\downarrow$ &11.25°$\uparrow$&22.5°$\uparrow$& 30°$\uparrow $&mean$\downarrow$ &med$\downarrow$ &11.25°$\uparrow$&22.5°$\uparrow$& 30°$\uparrow $\\
   \midrule
   Projection~\cite{Wang2025a}&
  75.06&80.31&0.066&0.124&0.156& 67.23&69.79&0.072&0.149&0.197 \\
   PriorDA~\cite{wang2025depthprior}&
   48.00&36.29&0.097&0.261&0.363 &36.26&30.30&0.181&0.405&0.519 \\
    CUT3R~\cite{Wang2025}&
   39.93&32.83&0.168&0.371&0.474 &34.64&26.20&0.219&0.447&0.548 \\
   FLARE~\cite{Zhang2025d}
   &52.29 &46.31 &0.111 &0.259 &0.346    &33.25 &24.63 &0.255 &0.471 &0.579 \\
    RayletDF~\cite{Wei2025}&
   39.87&34.70&0.084&0.288&0.439 &28.96&24.57&0.150&0.451&0.630\\
       Ours \textit{w/o refine}&
  36.49&28.29&0.193&0.380&0.488 &40.47&38.56&0.061&0.220&0.353\\
   \textbf{Ours} &
\textbf{29.83}&\textbf{21.21}&\textbf{0.292}&\textbf{0.557}&\textbf{0.654}& \textbf{20.72}&\textbf{12.67}&\textbf{0.476}&\textbf{0.711}&\textbf{0.781}\\
\bottomrule
\end{tabular}}
\label{tab:exp_ray_surface_normal_esti} \vspace{-0.5cm}
\end{table}

\subsection{Influence of the Number of Input Views}
We evaluate the effect of input-view density by varying the number of input views from 5 to 100. As shown in Tables~\ref{tab:exp_ray_surface_dis_sparse}\&~\ref{tab:exp_ray_surface_dis_dense}, \nickname{} consistently achieves superior reconstruction accuracy across all settings. Its advantage is particularly pronounced given sparse views, demonstrating stronger robustness under limited observations. Additional qualitative results are provided in Appendix~\ref{app:abl_views}.

\begin{table}[t]\tabcolsep=0.12cm 
\centering
\captionsetup{skip=2pt}
\caption{Quantitative results of estimated ray-surface distances under sparse views on ScanNet++.}
  \resizebox{1\textwidth}{!}{
\begin{tabular}{@{}r|ccccc|ccccc|ccccc@{}}
    \toprule
   &\multicolumn{5}{c|}{5 Views}&\multicolumn{5}{c|}{10 Views}&\multicolumn{5}{c}{25 Views}\\
   \cmidrule{2-16}
   &ADE$\downarrow$ &RMSE$\downarrow$ &Abs-Rel$\downarrow$&Sq-Rel$\downarrow$&$\delta\uparrow $&ADE$\downarrow$&RMSE$\downarrow$&Abs-Rel$\downarrow$&Sq-Rel$\downarrow$&$\delta\uparrow$&
   ADE$\downarrow$&RMSE$\downarrow$&Abs-Rel$\downarrow$&Sq-Rel$\downarrow$&$\delta\uparrow$ \\
   \midrule
   Projection~\cite{Wang2025a}&
   1.604&1.812&0.884&1.618&0.117 & 1.364&1.650&0.773&1.386&0.234  &0.927&1.302&0.551&0.923&0.450 \\
   PriorDA~\cite{wang2025depthprior}&
   0.635&0.776&0.397&0.583&0.565 &0.395&0.525&0.252&0.338&0.729
   &0.256&0.360&0.164&0.168&0.822\\
    CUT3R~\cite{Wang2025}&
   0.484&0.616&0.306&0.352&0.589 &0.406&0.524&0.262&0.251&0.642    &0.303&0.413&0.199&0.146&0.736\\
   FLARE~\cite{Zhang2025d}&
   1.164 &1.456 &0.615 &1.114 &0.353 &0.816 &1.126 &0.438 &0.731 &0.512   &0.448 &0.702 &0.234 &0.326 &0.708\\
    RayletDF~\cite{Wei2025}&
   0.679&0.920&0.440&1.001&0.633 &0.417&0.622&0.270&0.613&0.763    &0.231&0.396&0.149&0.285&0.872\\
   \textbf{Ours} &
   \textbf{0.370}&\textbf{0.514}&\textbf{0.253}&\textbf{0.260}&\textbf{0.690}& \textbf{0.252}&\textbf{0.394}&\textbf{0.164}&\textbf{0.168}&\textbf{0.800} &\textbf{0.166}&\textbf{0.300}&\textbf{0.104}&\textbf{0.093}&\textbf{0.878}\\
\bottomrule
\end{tabular}}
\label{tab:exp_ray_surface_dis_sparse} \vspace{-0.5cm}
\end{table}

\begin{table}[!htbp]\tabcolsep=0.12cm 
\centering
\captionsetup{skip=2pt}
\caption{Quantitative results of estimated ray-surface distances under dense views on ScanNet++.}
  \resizebox{1\textwidth}{!}{
\begin{tabular}{@{}r|ccccc|ccccc|ccccc@{}}
    \toprule
   &\multicolumn{5}{c|}{50 Views}&\multicolumn{5}{c|}{70 Views}&\multicolumn{5}{c}{100 Views}\\
   \cmidrule{2-16}
   &ADE$\downarrow$ &RMSE$\downarrow$ &Abs-Rel$\downarrow$&Sq-Rel$\downarrow$&$\delta\uparrow $&ADE$\downarrow$&RMSE$\downarrow$&Abs-Rel$\downarrow$&Sq-Rel$\downarrow$&$\delta\uparrow$&
   ADE$\downarrow$&RMSE$\downarrow$&Abs-Rel$\downarrow$&Sq-Rel$\downarrow$&$\delta\uparrow$ \\
   \midrule
   Projection~\cite{Wang2025a}&
   0.580&0.959&0.360&0.554&0.636    & 0.469&0.831&0.294&0.434&0.696  &0.363&0.691&0.226&0.317&0.762 \\
   PriorDA~\cite{wang2025depthprior}&
   0.198&0.290&0.123&0.101&0.868         &0.190&0.283&0.116&0.086&0.874
   &0.180&0.278&0.105&0.077&0.887\\
    CUT3R~\cite{Wang2025}&
   0.335&0.446&0.222&0.173&0.703 &0.359&0.470&0.239&0.181&0.667    &0.395&0.507&0.266&0.208&0.623\\
   FLARE~\cite{Zhang2025d}&
   0.312&0.495&0.157&0.156&0.771 &0.288 &0.458 &0.143 &0.123 &0.784   &0.281 &0.464 &0.139 &0.128 &0.796\\
    RayletDF~\cite{Wei2025}&
   0.179&0.334&0.119&0.205&0.896 &0.170&0.322&0.111&0.159&0.893    &0.162&0.617&0.107&0.138&0.896\\
   \textbf{Ours} &
   \textbf{0.132}&\textbf{0.259}&\textbf{0.079}&\textbf{0.067}&\textbf{0.910}& \textbf{0.126}&\textbf{0.248}&\textbf{0.076}&\textbf{0.062}&\textbf{0.916} &\textbf{0.121}&\textbf{0.240}&\textbf{0.072}&\textbf{0.059}&\textbf{0.922}\\
\bottomrule
\end{tabular}}
\label{tab:exp_ray_surface_dis_dense} \vspace{-0.5cm}
\end{table}

\subsection{Evaluation of Single View Reconstruction}
In this section, we evaluate single-view reconstruction. For each test sequence, we sample one image as the input view and use its two adjacent frames as query views for evaluation. Quantitative results are reported in Table~\ref{tab:exp_ray_surface_dis_esti_single}. Our method still achieves the best performance across all datasets. CUT3R obtains the second-best results on most datasets, likely benefiting from its training design for unseen-view inpainting. Additional qualitative results are provided in Appendix~\ref{app:single_view}.

\begin{table}[!htbp]\tabcolsep=0.12cm 
\centering
\captionsetup{skip=2pt}
\caption{Quantitative results of single-view reconstruction.}
  \resizebox{1\textwidth}{!}{
\begin{tabular}{@{}r|ccccc|ccccc|ccccc@{}}
    \toprule
   &\multicolumn{5}{c|}{\textit{test on} $\rightarrow$ ScanNet++}&\multicolumn{5}{c|}{\makecell{\textit{test on} $\rightarrow$ 7Scenes}}&\multicolumn{5}{c}{\makecell{\textit{test on} $\rightarrow$ MultiScan}}\\
   \cmidrule{2-16}
   &ADE$\downarrow$ &RMSE$\downarrow$ &Abs-Rel$\downarrow$&Sq-Rel$\downarrow$&$\delta\uparrow $&ADE$\downarrow$&RMSE$\downarrow$&Abs-Rel$\downarrow$&Sq-Rel$\downarrow$&$\delta\uparrow$&
   ADE$\downarrow$&RMSE$\downarrow$&Abs-Rel$\downarrow$&Sq-Rel$\downarrow$&$\delta\uparrow$ \\
   \midrule
   Projection~\cite{Wang2025a}&
   1.454&1.759&0.766&1.444&0.233& 0.910&1.323&0.467&0.848&0.533&0.517&0.807&0.398&0.479&0.602 \\
   PriorDA~\cite{wang2025depthprior}&
   0.174&0.288&0.096&0.092&0.895 &0.155&0.247&0.076&\textbf{0.031}&0.924    &0.172&0.275&0.138&0.082&0.820\\
    CUT3R~\cite{Wang2025}&
   0.139&0.249&0.079&0.047&0.922 &0.153&0.249&0.077&0.037&0.925    &0.151&0.225&0.130&0.083&0.852\\
   FLARE~\cite{Zhang2025d}
   &0.301 &0.559 &0.154 &0.203 &0.835   &0.257 &0.497 &0.127 &0.145 &0.866    &0.265 &0.434 &0.254 &0.221 &0.781\\
    RayletDF~\cite{Wei2025}&
   0.150&0.280&0.129&0.117&0.869 &0.162&0.277&0.085&0.046&0.908    &0.212&0.341&0.193&0.155&0.775\\
   \textbf{Ours} &
   \textbf{0.100}&\textbf{0.221}&\textbf{0.061}&\textbf{0.044}&\textbf{0.930}& \textbf{0.134}&\textbf{0.230}&\textbf{0.068}&\textbf{0.031}&\textbf{0.941} &\textbf{0.122}&\textbf{0.209}&\textbf{0.108}&\textbf{0.056}&\textbf{0.882}\\
   \midrule
    &\multicolumn{5}{c|}{\textit{test on} $\rightarrow$ Replica}&\multicolumn{5}{c|}{\makecell{\textit{test on} $\rightarrow$ MuSHRoom}}&\multicolumn{5}{c}{\makecell{\textit{test on} $\rightarrow$ Tanks\&Temples}}\\
   \cmidrule{2-16}
    Projection~\cite{Wang2025a}&
    2.466&2.768&0.863&2.459&0.137 &1.690&1.957&0.905&1.685&0.099 &9.517&11.87&0.895&13.40&0.241\\
   PriorDA~\cite{wang2025depthprior}&
   0.219&0.365&0.087&0.648&0.885 &0.359&0.550&0.213&0.209&0.725 &1.619&\textbf{2.828}&0.198&1.163&0.802\\
    CUT3R~\cite{Wang2025}&
   0.230&0.363&0.089&0.059&0.920 &0.245&0.401&0.170&0.149&0.825    &1.919&3.075&\textbf{0.146}&\textbf{1.137}&0.833\\
   FLARE~\cite{Zhang2025d}
   &0.901 &1.536 &0.313 &0.795 &0.689   &0.758 &1.220 &0.405 &0.676 &0.559    &3.830 &5.995 &0.254 &3.102 &0.709\\
    RayletDF~\cite{Wei2025}&
   0.236&0.496&0.105&0.162&0.882 &0.357&0.599&0.298&0.486&0.772    &3.034&6.299&0.548&14.79&0.774\\
   \textbf{Ours}  & 
   \textbf{0.121}&\textbf{0.252}&\textbf{0.050}&\textbf{0.034}&\textbf{0.952} &\textbf{0.185}&\textbf{0.347}&\textbf{0.141}&\textbf{0.131}&\textbf{0.872 }&\textbf{1.480}&3.386&{0.309}&6.846&\textbf{0.899}\\
\bottomrule
\end{tabular}}
\label{tab:exp_ray_surface_dis_esti_single}\vspace{-0.5cm}
\end{table}

\subsection{Ablation Study}
We investigate the effectiveness of surfel matching, hierarchical features, local coordinate, and reference view selection. Due to computational constraints, as full training requires approximately 288 A800 GPU hours, we train the ablation variants on ScanNet and evaluate them on the test split of MuSHRoom dataset. Additional results are provided in Appendix~\ref{app:scannet_abl}.

\textbf{Effects of Surfel Voting.}
For reference-view selection, we compare two strategies: (1) estimating view overlap via per-pixel comparison, and (2) our surfel voting strategy. As shown in Table~\ref{tab:exp_abl_mushroom}, surfel voting consistently improves performance, indicating that surfel-level aggregation provides a more reliable measure of cross-view relevance. Additional comparisons are provided in Appendix~\ref{app:surfel_abl}.

\textbf{Effects of Hierarchical Features.}
To evaluate the impact of hierarchical features, we ablate our model to use only the final-layer VGGT tokens. As Table~\ref{tab:exp_abl_mushroom} shows, this removal increases all error metrics, highlighting that multi-scale representations are vital for capturing the global structure and local details required for accurate ray-surface distance regression.

\textbf{Effects of Local Coordinates.}
We evaluate the choice of coordinate space by comparing global VGGT coordinates against a query-view-centric local coordinates. Table~\ref{tab:exp_abl_mushroom} shows that adopting local coordinates provides distinct performance gains, highlighting its stronger generalization.

\textbf{Effects of Reference-view Selection.}
As shown in the last row of Table~\ref{tab:exp_abl_mushroom}, removing the \textit{top-k} reference-view selection severely degrades performance. This confirms that querying all views without filtering introduces detrimental noise and irrelevant observations.

\begin{table}[!htbp] 
\tabcolsep=0.15cm  
\centering
\captionsetup{skip=2pt}
\caption{Ablation results on the test split of MuSHRoom.}
\label{tab:exp_abl_mushroom}
\resizebox{0.6\linewidth}{!}
{
\begin{tabular}{lccccc}
\toprule[1.0pt]
& ADE$\downarrow$ &RMSE$\downarrow$ &Abs-Rel$\downarrow$&Sq-Rel$\downarrow$&$\delta\uparrow $  \\
\toprule[1.0pt]
w/o surfel vote     &0.341&0.550&0.278&0.401&0.754\\
w/o hierarchy feature     &0.369&0.575&0.307&0.444&0.732\\
w/o local coordinate   &0.457 &0.658&0.367&0.628 &0.685\\
w/o \textit{top-k} reference view &0.372&0.577&0.285&0.375&0.722\\
Full \nickname{}    &\textbf{0.331}&\textbf{0.538}&\textbf{0.266}&\textbf{0.372}&\textbf{0.761}\\
\toprule[1.0pt]
\end{tabular}
}\vspace{-0.4cm}
\end{table}

\subsection{Sensitivity to Geometric Priors}
Since FoundSurface builds upon geometric priors predicted by VGGT, we further investigate its sensitivity to errors in the estimated pointmaps and camera poses. Specifically, we group test samples according to their VGGT pointmap errors and camera pose errors, measured by Absolute Distance Error (ADE) and Absolute Trajectory Error (ATE), respectively. As shown in the following Table ~\ref{tab:vggt_pointmap_error} and ~\ref{tab:vggt_camera_error}, our method maintains a consistent error level across different VGGT results, demonstrating its robustness. 

\begin{table}[!htbp]
    \centering
    \begin{minipage}[t]{0.495\linewidth}
        \centering
        \caption{Results across different VGGT pointmap error levels on ScanNet++.}
        \label{tab:vggt_pointmap_error}
        \vspace{2pt}
        \footnotesize
        \setlength{\tabcolsep}{1.5pt}
        \begin{tabular}{@{}lccccc@{}}
            \toprule
            VGGT ADE & ADE$\downarrow$ & RMSE$\downarrow$ &
            Abs-Rel$\downarrow$ & Sq-Rel$\downarrow$ & $\delta\uparrow$ \\
            \midrule
            0--0.05m    & 0.041 & 0.107 & 0.032 & 0.012 & 0.973 \\
            0.05--0.10m & 0.113 & 0.250 & 0.070 & 0.051 & 0.928 \\
            0.10--0.20m & 0.167 & 0.324 & 0.107 & 0.080 & 0.875 \\
            $>$0.20m    & 0.534 & 0.765 & 0.265 & 0.350 & 0.617 \\
            \bottomrule
        \end{tabular}
    \end{minipage}
    \hfill
    \begin{minipage}[t]{0.495\linewidth}
        \centering
        \caption{Results across different VGGT camera error levels on ScanNet++.}
        \label{tab:vggt_camera_error}
        \vspace{2pt}
        \footnotesize
        \setlength{\tabcolsep}{1.5pt}
        \begin{tabular}{@{}lccccc@{}}
            \toprule
            VGGT ATE & ADE$\downarrow$ & RMSE$\downarrow$ &
            Abs-Rel$\downarrow$ & Sq-Rel$\downarrow$ & $\delta\uparrow$ \\
            \midrule
            0--0.05m    & 0.069 & 0.172 & 0.044 & 0.027 & 0.958 \\
            0.05--0.10m & 0.123 & 0.260 & 0.079 & 0.058 & 0.916 \\
            0.10--0.20m & 0.153 & 0.287 & 0.099 & 0.071 & 0.895 \\
            $>$0.20m    & 0.444 & 0.645 & 0.227 & 0.678 & 0.799 \\
            \bottomrule
        \end{tabular}
    \end{minipage}
\end{table}

\subsection{Evaluation of Surface Reconstruction with 3D Metrics}
\label{sec:surface_eval}

We further evaluate the reconstructed 3D geometry on ScanNet++ using surface completeness, Chamfer distance (CD), accuracy, and F-scores. For a controlled comparison, we apply TSDF, Poisson, and NKSR to identical VGGT pointmaps. We also compare with generalizable feed-forward methods and PGSR, a per-scene optimization baseline. For all remaining methods, we reconstruct meshes by TSDF fusion of their predicted novel-view and VGGT points.

\begin{table}[!htbp]
\vspace{-0.4cm}
\centering
\captionsetup{skip=2pt}
\caption{Scores of 3D metrics for surface reconstruction metrics on ScanNet++.}
\label{tab:surface_comparison}
\setlength{\tabcolsep}{4pt}
\scriptsize
\renewcommand{\arraystretch}{0.9}
\begin{tabular*}{\textwidth}{@{\extracolsep{\fill}}llccccc@{}}
\toprule
Category & Method & Comp.$\downarrow$ & CD$\downarrow$ &
Acc.$\downarrow$ & F1@5cm$\uparrow$ & F1@10cm$\uparrow$ \\
\midrule
\multirow{3}{*}{Surface Fitting}
& VGGT + TSDF~\cite{curless1996volumetric}    & 0.148 & 0.095 & \textbf{0.042} & 0.724 & 0.847 \\
& VGGT + Poisson~\cite{kazhdan2013screened} & 0.102 & 0.074 & 0.047 & 0.787 & 0.897 \\
& VGGT + NKSR~\cite{huang2023neural}    & 0.166 & 0.124 & 0.082 & 0.647 & 0.758 \\
\midrule
\multirow{3}{*}{\makecell[l]{Generalizable Feed-forward\\Reconstruction}}
& PriorDA~\cite{wang2025depthprior}  & 0.108 & 0.080 & 0.052 & 0.721 & 0.873 \\
& CUT3R~\cite{Wang2025}    & 0.205 & 0.191 & 0.177 & 0.212 & 0.423 \\
& FLARE~\cite{Zhang2025d}    & 0.122 & 0.131 & 0.140 & 0.440 & 0.704 \\
\midrule
Per-scene Optimization
& PGSR~\cite{Chen2024a} & 0.081 & 0.090 & 0.100 & 0.616 & 0.779 \\
\midrule
\multirow{4}{*}{\makecell[l]{Feed-forward\\Ray-based Model}}
& RayletDF~\cite{Wei2025} & 0.076 & 0.088 & 0.099 & 0.702 & 0.838 \\
&Ours w/o refine & 0.078 & 0.064 & 0.049 & 0.729 & 0.879 \\
&Ours (voxel 0.025) & 0.091 & 0.066 & \textbf{0.042} & 0.761 & 0.883 \\
& \textbf{Ours (voxel 0.01)}
& \textbf{0.066} & \textbf{0.056} & 0.046
& \textbf{0.792} & \textbf{0.910} \\
\bottomrule
\end{tabular*}
\vspace{-0.2cm}
\end{table}

\textbf{Results \& Analysis.}
As shown in Table~\ref{tab:surface_comparison}, our method achieves the best completeness, CD, and F-scores, while maintaining competitive accuracy. Notably, it outperforms classical surface fitting methods using identical VGGT pointmaps, demonstrating the advantage of our learned ray-based surface representation. The results also confirm the effectiveness of raylet-based refinement in improving 3D surface reconstruction quality.

\section{Conclusion}\label{sec:conclusion}

In this paper, we introduced a generalizable, feed-forward model that reconstructs continuous, high-fidelity 3D surfaces from unposed image collections. By seamlessly integrating pointmap- and ray-based representations through a query-view-centric selection module and a raylet-based refinement network, our method bridges the gap between discrete geometric priors and continuous surface generation. Extensive cross-dataset evaluations demonstrate that our approach achieves superior zero-shot generalization and geometric accuracy compared to existing baselines.

\textbf{Limitations}: Our model has a large size due to the reliance on a pre-trained pointmap backbone. Future work will focus on optimizing the network architecture for deployment on mobile devices.

\clearpage

\textbf{Acknowledgments}:
This work was supported in part by Research Grants Council of Hong Kong under Grants 15225522 \&15228626 \& 15219125, in part by National Natural Science Foundation of China under Grant 62271431, in part by Otto Poon Charitable Foundation Smart Cities Research Institute (8-CDCQ), in part by Research Center for Unmanned Autonomous Systems (1-CE3D), and in part by PolyU Kunpeng \& Ascend Technology Innovation Incubation Center, The Hong Kong Polytechnic University.  

\bibliographystyle{abbrv}
\bibliography{ref.bib}


\appendix

\clearpage
\section{Appendix}
\label{sec:appendix}

\noindent\textbf{The appendix includes:}

\begin{itemize}
    \item Details for query-view-centric selection.
    \item Details of coarse ray-surface distance regressor.
    \item Details of raylet-based surface module.
    \item Details of training.
    \item Computation cost analysis.
    \item Results of standard deviation.
    \item The definitions of all the evaluation metrics. 
    \item More quantitative and qualitative results.
    \item Details of datasets.
    \item Implementation details for our model.
    \item Robustness evaluation of the S-nearest neighbor.
    \item Qualitative results of 3D outputs.
    \item Failure cases.
    \item Effectiveness of raylet-Based refinement.
    \item Efficiency of ray-Based surface queries.\\
\end{itemize}

\subsection{Details of Query-View-Centric Selection}
\label{app:detail_ref_selection}
For reference view selection, we first project the $N$ point coordinate frames to the query camera plane using the extrinsics and intrinsics predicted by VGGT, resulting in $N\times H\times W$ projected images. A pseudo-depth map for the query view is then generated by taking the minimum positive value across these projections. Lastly, both the projected images and the pseudo-depth map are divided into 8×8 non-overlapping patches. The candidate view evaluation uses thresholds of $\delta_{depth}$=0.05 and $\tau_{vote}$=0.3. For all experimental evaluations, the number of selected reference views is fixed at 2.

\subsection{Details of Coarse Ray-Surface Distance Regressor}
\label{app:detail_coarse_regressor}
\subsubsection{Details of Ray Embedding Generation}
Given the camera parameters $\textbf{T}_{target}$ and $\textbf{C}_{target}$, ray parameter of pixel $(u,v)$ is computed by
\begin{equation}
     o = \textbf{T}_{target}[:3,3] 
\end{equation}
\begin{equation}
d = \text{Normalize}(\textbf{T}_{target}\cdot\textbf{C}_{target}^{-1}\cdot[u,v,1])
\end{equation}
Then the Plücker parameter of $(o,d)$ is calculated by: $(d, \ o\times d)$.
Subsequently, the dense ray map is tokenized via a convolutional layer with a kernel and stride of 14.

\subsubsection{Details of Spatial Feature Aggregation Transformer}
The aggregation module, denoted as $f_{\text{enc}}$, is implemented as a Transformer-based architecture comprising four alternating blocks of self-attention and cross-attention layers. To balance computational efficiency and modeling capacity, each attention layer is configured with four heads. This structure facilitates the effective fusion of global contextual information with localized ray-specific features.

\subsection{Details of Raylet-based Surface Module}
\label{app_normal_dev}
\subsubsection{Details of Raylet Network}
The Raylet refinement network $f_{\text{refine}}$ consists of three modules: (1) a dimensionality reduction network that contains two linear layers and a ReLU layer. (2) a Raylet prediction network which is an 8-layer MLP network. (3) a sparse convolution network to aggregate local features in 3D space. Considering that the pointmaps estimated by VGGT are defined within a normalized coordinate space, we set the voxelization resolution to $0.025$. This configuration inherently facilitates an adaptive voxelization strategy, where the effective resolution scales dynamically according to the scene's spatial extent.
Structure of Raylet prediction network and sparse convolution network is available in Section~\ref{app_spconv}.

\subsubsection{Surface Normal Derivation}
Similar to RayletDF, our method also has a nice property of deriving a closed-form expression for a normal vector, thanks to the nature of our ray-based formulation. Given a ray from camera center, we can  parameter its direction in spherical coordinate system as $(\theta, \phi, 1)$ and convert the ray direction back to world coordinate as $\bm{r} = (x,y,z) = (\sin{\theta}\cos{\phi}, \cos{\theta}, \sin{\theta}\sin{\phi})$. Then we utilize the coarse ray-surface distance regressor to compute the initial intersection ray distance $D^{target}$. Consequently, we construct a raylet $\bm{l}$ and calculate the complete ray distance by $D = D^{target} + \Delta D$; the final estimated surface point can be formed as:
\begin{equation}
     \Phi(\theta, \phi) = (D\sin{\theta}\cos{\phi}, D\cos{\theta}, D\sin{\theta}\sin{\phi})
     \label{surface_pts}
 \end{equation}

Since the full network is differentiable, we can calculate the derivative of the surface point with respect to $\theta, \phi$ by:

\begin{equation}
\frac{\partial \Phi}{\partial \phi} = 
\begin{bmatrix}
\left( \frac{\partial D}{\partial \phi} \cos \phi - D \sin \phi \right) \sin \theta \\
\frac{\partial D}{\partial \phi} \cos \theta\\
\left( \frac{\partial D}{\partial \phi} \sin \phi + D \cos \phi \right) \sin \theta 
\end{bmatrix}
\end{equation}

\begin{equation}
\frac{\partial \Phi}{\partial \theta} = 
\begin{bmatrix}
\left( \frac{\partial D}{\partial \theta} \sin \theta + D \cos \theta \right) \cos \phi \\
\frac{\partial D}{\partial \theta} \cos \theta - D \sin \theta\\
\left( \frac{\partial D}{\partial \theta} \sin \theta + D \cos \theta \right) \sin \phi \\

\end{bmatrix}
\end{equation}
Lastly, the formula for a unit normal vector is :
\begin{equation}
    \mathbf{n} = \frac{\frac{\partial \Phi}{\partial \phi} \times \frac{\partial \Phi}{\partial \theta}}{\left\| \frac{\partial \Phi}{\partial \phi} \times \frac{\partial \Phi}{\partial \theta} \right\|}.
\end{equation}

\subsection{Details of Training}
\label{app_train}
The VGGT prediction and ground truth are not located in the same coordinate system; thus, we need to align novel view camera pose for supervision. Though the VGGT is expected to predict at a canonical normalized scale during training stage, in inference, it does not perfectly align with it. Directly training the network to predict novel views on data normalized by VGGT preprocessing pipeline would force the network to compensate for the coordinate system mismatch. Thus, we re-calculate the transformation from VGGT prediction to ground truth data during training. 
Specifically, we first filter out the points whose confidence scores are within the top $85\%$. Then least-square fit algorithm~\cite{bregier2021deepregression} is utilized to compute the scale, rotation and translation between the VGGT prediction to ground truth data, forming a transformation matrix. Subsequently, we use the transformation matrix to align the ground truth data for supervision. 

Our models are trained using the AdamW optimizer and a OneCycle learning rate schedule. Specifically, the coarse regressor is trained for 200 epochs with a learning rate of $1\times10^{-4}$, while the refinement network is trained for 100 epochs with a learning rate of $1\times10^{-4}$. In each epoch, we sample 4,200 data points for training. The model is trained on 4 A800 GPUs for about 72 hours.

\subsection{Computation Cost Analysis}

We report the memory consumption and rendering speed in test for PriorDA, CUT3R, FLARE, RayletDF and ours on ScanNet test split with 50 input views in Table \ref{tab:speed}. All methods are tested on a single A800 GPU with Intel(R) Xeon(R) Platinum 8358 CPU. Voxel grid KNN \url{https://github.com/janericlenssen/torch_knnquery/} is employed to accelerate our algorithm. Since our method only needs 1 raylet sampled on each ray and each raylet only has 5 neighboring points sampled, for a fair comparison we also only sample 1 raylet for RayletDF. We can see that: 1) our method has a clear advantage over the depth completion method in rendering speed; 2) though being slower than CUTR3R and FLARE, our method achieves much higher accuracy in surface reconstruction.  

\begin{table*}[h] 
\centering
 \caption{{The time, memory consumption, and the speed of rendering a $640\times480$ image of ScanNet.}}\label{tab:speed}
\resizebox{0.7\linewidth}{!}{
\begin{tabular}{r|cccc}
\toprule[1.0pt]
& Memory (GB) & No. of Samples per ray & Render Speed (FPS) & ADE $\downarrow$\\
\toprule[1.0pt]
    PiorDA~\cite{wang2025depthprior}  & 17.1&     - &1.55 & 0.125 \\ 
    CUT3R~\cite{Wang2025}  & 8.9&     - &16.7& 0.298 \\ 
    FLARE~\cite{Zhang2025d}  &19.4 & - &31.9 & 0.238 \\ 
    RayletDF (5 samples)~\cite{Wei2025}  &11.2 & 5$\times$5 &6.74 & 0.136 \\ 
    RayletDF (1 sample)~\cite{Wei2025}  &3.17 & 1$\times$5 &26.2 & 0.202 \\ 
    Ours   &  20.89&     1$\times$5 &8.68 & 0.089 \\ 
    \hline
    \toprule[1.0pt]
  \end{tabular}
}
\end{table*}

We further evaluate the computational efficiency of our method under varying image resolutions and numbers of input views. As shown in Tables~\ref{tab:mem_res} and~\ref{tab:mem_view}, our method maintains moderate GPU memory consumption and supports high-resolution rendering and larger input-view sets with manageable computational overhead.

\begin{table}[!h] 
\tabcolsep=0.15cm  
\centering
\caption{The memory consumption and rendering speed for varying image resolutions.}
\label{tab:mem_res}
\resizebox{0.6\linewidth}{!}
{
\begin{tabular}{ccc}
\toprule[1.0pt]
Resolution& Render Speed (FPS) & GPU Memory (GB)  \\
\toprule[1.0pt]
320*240    &18.49 & 20.69 \\
640*480    &8.68 & 20.89 \\
960*720  &4.59 & 21.27\\
1440*1080     &2.13& 22.18\\
\toprule[1.0pt]
\end{tabular}
}
\end{table}

\subsection{Standard Deviation Results}
\label{app:std}
We report standard deviation results on out-of-domain datasets in Table \ref{tab:exp_ray_surface_dis_esti_std}. As shown in the Table, our method achieves the most robust performance.

\begin{table*}[ht]\tabcolsep=0.12cm 
\centering
\caption{Quantitative results of standard derivation.}
  \resizebox{1\textwidth}{!}{
\begin{tabular}{@{}r|ccccc|ccccc|ccccc@{}}
    \toprule
   &\multicolumn{5}{c|}{\textit{test on} $\rightarrow$ ScanNet++}&\multicolumn{5}{c|}{\makecell{\textit{test on} $\rightarrow$ 7Scenes}}&\multicolumn{5}{c}{\makecell{\textit{test on} $\rightarrow$ MultiScan}}\\
   \cmidrule{2-16}
   &ADE$\downarrow$ &RMSE$\downarrow$ &Abs-Rel$\downarrow$&Sq-Rel$\downarrow$&$\delta\uparrow $&ADE$\downarrow$&RMSE$\downarrow$&Abs-Rel$\downarrow$&Sq-Rel$\downarrow$&$\delta\uparrow$&
   ADE$\downarrow$&RMSE$\downarrow$&Abs-Rel$\downarrow$&Sq-Rel$\downarrow$&$\delta\uparrow$ \\
   \midrule
   Projection~\cite{Wang2025a}&
    $\pm$0.129&$\pm$0.170&$\pm$0.103&$\pm$0.134&$\pm$0.127& $\pm$0.073&$\pm$0.155&$\pm$0.032&$\pm$0.056&$\pm$0.042&$\pm$0.287&$\pm$0.302&$\pm$0.211&$\pm$0.327&$\pm$0.190 \\
   PriorDA~\cite{wang2025depthprior}&
   $\pm$0.129&$\pm$0.170&$\pm$0.103&$\pm$0.134&$\pm$0.127 &$\pm$0.047&$\pm$0.082&$\pm$0.183&$\pm$0.017&$\pm$0.042   &$\pm$0.268&$\pm$0.294&$\pm$0.210&$\pm$0.319&$\pm$0.205\\
    CUT3R~\cite{Wang2025}&
   $\pm$0.146&$\pm$0.190&$\pm$0.135&$\pm$0.200&$\pm$0.200 &$\pm$0.045&$\pm$0.092&$\pm$0.025&$\pm$0.028&$\pm$0.040    &$\pm$0.209&$\pm$0.227&$\pm$0.352&$\pm$0.493&$\pm$0.251\\
   FLARE~\cite{Zhang2025d}
   &$\pm$0.334 &$\pm$0.416 &$\pm$0.136 &$\pm$0.293 &$\pm$0.206   &$\pm$0.103 &$\pm$0.174 &$\pm$0.040 &$\pm$0.073 &$\pm$0.090  &$\pm$0.411 &$\pm$0.454 &$\pm$0.273 &$\pm$0.432 &$\pm$0.276\\
    RayletDF~\cite{Wei2025}&
   $\pm$0.145&$\pm$0.229&$\pm$0.121&$\pm$0.385&$\pm$0.080 &$\pm$0.045&$\pm$0.092&$\pm$0.025&$\pm$0.028&$\pm$0.040       &$\pm$0.251&$\pm$0.308&$\pm$0.276&$\pm$0.671&$\pm$0.165  \\
     Ours \textit{w/o refine}&
   $\pm$0.093&\textbf{$\pm$0.151}&$\pm$0.065&\textbf{$\pm$0.080}&$\pm$0.097 &$\pm$0.036&\textbf{$\pm$0.070}&$\pm$0.019&\textbf{$\pm$0.017}&\textbf{$\pm$0.026 }   &$\pm$0.131&$\pm$0.168&$\pm$0.179&$\pm$0.207&$\pm$0.203\\
   \textbf{Ours} &
   \textbf{$\pm$0.087}&$\pm$0.155&\textbf{$\pm$0.060}&$\pm$0.087&\textbf{$\pm$0.079}& \textbf{$\pm$0.033}&$\pm$0.073&\textbf{$\pm$0.018}&$\pm$0.018&\textbf{$\pm$0.026} &\textbf{$\pm$0.121}&\textbf{$\pm$0.165}&\textbf{$\pm$0.158}&\textbf{$\pm$0.204}&\textbf{$\pm$0.160}\\
   \midrule
    &\multicolumn{5}{c|}{\textit{test on} $\rightarrow$ Replica}&\multicolumn{5}{c|}{\makecell{\textit{test on} $\rightarrow$ MuSHRoom}}&\multicolumn{5}{c}{\makecell{\textit{test on} $\rightarrow$ Tanks\&Temples}}\\
   \cmidrule{2-16}
    Projection~\cite{Wang2025a}&
    $\pm$0.146&$\pm$0.203&$\pm$0.076&$\pm$0.154&$\pm$0.077 &$\pm$0.304&$\pm$0.422&$\pm$0.374&$\pm$1.333&$\pm$0.165 &$\pm$1.507&$\pm$2.414&$\pm$0.219&$\pm$5.519&$\pm$0.126\\
   PriorDA~\cite{wang2025depthprior}&
   $\pm$0.046&$\pm$0.069&$\pm$0.029&$\pm$0.021&$\pm$0.037 &$\pm$0.254&$\pm$0.338&$\pm$0.421&$\pm$1.080&$\pm$0.204 &$\pm$1.022&\textbf{$\pm$1.638}&$\pm$0.125&\textbf{$\pm$1.326}&$\pm$0.144\\
    CUT3R~\cite{Wang2025}&
   $\pm$0.119&$\pm$0.140&$\pm$0.071&$\pm$0.082&$\pm$0.129 &$\pm$0.363&$\pm$0.431&$\pm$0.641&$\pm$1.641&$\pm$0.259   &$\pm$2.990&$\pm$3.838&$\pm$0.491&$\pm$11.83&$\pm$0.247\\
   FLARE~\cite{Zhang2025d}
   &$\pm$0.124 &$\pm$0.191 &$\pm$0.051 &$\pm$0.063 &$\pm$0.126   &$\pm$0.456 &$\pm$0.576 &$\pm$0.296 &$\pm$0.822 &$\pm$0.254    &$\pm$2.209 &$\pm$3.710 &$\pm$0.144 &$\pm$2.935 &$\pm$0.171\\
    RayletDF~\cite{Wei2025}&
  $\pm$0.033&$\pm$0.079&$\pm$0.024&$\pm$0.026&$\pm$0.023 &$\pm$0.350&$\pm$0.504&$\pm$0.587&$\pm$2.258&$\pm$0.177    &$\pm$1.773&$\pm$2.670&$\pm$0.343&$\pm$10.59&$\pm$0.124\\
     Ours \textit{w/o refine}&
   $\pm$0.037&\textbf{$\pm$0.058}&$\pm$0.020&\textbf{$\pm$0.014}&$\pm$0.024 &$\pm$0.227&\textbf{$\pm$0.332}&$\pm$0.374&\textbf{$\pm$1.087}&$\pm$0.213  &$\pm$1.132&$\pm$2.404&$\pm$0.129&$\pm$2.658&$\pm$0.129\\
   \textbf{Ours}  & 
   \textbf{$\pm$0.028}&$\pm$0.062&\textbf{$\pm$0.016}&\textbf{$\pm$0.014}&\textbf{$\pm$0.017} &\textbf{$\pm$0.213}&$\pm$0.354&\textbf{$\pm$0.359}&$\pm$1.228&\textbf{$\pm$0.166 }&\textbf{$\pm$0.950}&$\pm$2.370&\textbf{{$\pm$0.117}}&$\pm$2.743&\textbf{$\pm$0.082}\\
\bottomrule
\end{tabular}}
\label{tab:exp_ray_surface_dis_esti_std} \vspace{-0.3cm}
\end{table*}

\subsection{Ray-surface Distance Metrics}
\label{app_metric_ray}
We report ray distance metrics including Absolute Distance Error (ADE), RMSE, Absolute Relative Distance (Abs-Rel), Squared Relative Distance (Sq Rel) and Threshold accuracy ($\delta < t$) to measure the error between the predicted $i^{th}$ ray-surface distance $t_i^{pred}$ and the ground truth $t_i^{gt}$. These metrics are calculated over $N$ samples and defined as:

\noindent\textbf{Absolute Distance Error (ADE)}:
$$\frac{1}{N}\sum_{i=1}^N|t_i^{pred} - t_i^{gt}|$$
\noindent\textbf{RMSE}:
$$\sqrt{\frac{1}{N}\sum_{i=1}^N(t_i^{pred} - t_i^{gt})^2}$$
\noindent\textbf{Absolute Relative Distance (Abs Rel)}:
$$\frac{1}{N}\sum_{i=1}^N\frac{|t_i^{pred} - t_i^{gt}|}{t_i^{gt}}$$
\noindent\textbf{Squared Relative Distance (Sq Rel)}:
$$\frac{1}{N}\sum_{i=1}^N\frac{(t_i^{pred} - t_i^{gt})^2}{t_i^{gt}} $$
\noindent\textbf{Threshold accuracy, $\delta$}:
$$\frac{1}{N}\sum_{i=1}^N[\text{max}(\frac{t_i^{pred}}{t_i^{gt}}, \frac{t_i^{gt}}{t_i^{pred}})<\delta]$$

\noindent\textbf{Absolute Trajectory Error (ATE)}:
$$
\frac{1}{N}\sum_{i=1}^N \left\| \mathbf{c}_i^{pred} - \mathbf{c}_i^{gt} \right\|_2
$$
where \(\mathbf{c}_i^{pred},\mathbf{c}_i^{gt}\in\mathbb{R}^3\) denote the predicted and ground-truth 3D camera centers of the \(i\)-th frame, respectively.

\subsection{Evaluation Metrics on Normals}
\label{app_metric_norm}
Given the ground truth normal $\mathbf{n}_{g,i}$ and predicted normal $\mathbf{n}_{p,i}$, per-pixel angle errors are first computed by:
$$    \theta_i = \arccos(\mathbf{n}_{p,i} \cdot \mathbf{n}_{g,i})$$

Then metrics reported in our paper are calculated over N samples and defined as following:

\noindent\textbf{Mean Absolute Error (MAE)}:
$$   \text{MAE} = \frac{1}{N} \sum_{i=1}^{N} \theta_i$$

\noindent\textbf{Median Absolute Error (MAE)}:
$$\text{Median} = \text{median}(\{\theta_i\}_{i=1}^{N})$$

\noindent\textbf{Fraction of pixels where $\theta < t$}:

$$\delta < t = \frac{1}{N} \sum_{i=1}^{N} [ \theta_i < t ] \times 100\%$$

\subsection{Evaluation Metrics on Meshes}
\label{app_metric_mesh}

We uniformly sample points from the predicted and ground-truth meshes, denoted as $\hat{P}$ and $P$, respectively. The evaluation metrics are defined as follows:

\noindent\textbf{Accuracy}:
\[
\mathrm{Acc} =
\frac{1}{|\hat{P}|}
\sum_{\hat{\mathbf{p}}\in\hat{P}}
\min_{\mathbf{p}\in P}
\|\hat{\mathbf{p}}-\mathbf{p}\|_2.
\]

\noindent\textbf{Completeness}:
\[
\mathrm{Comp} =
\frac{1}{|P|}
\sum_{\mathbf{p}\in P}
\min_{\hat{\mathbf{p}}\in\hat{P}}
\|\mathbf{p}-\hat{\mathbf{p}}\|_2.
\]

\noindent\textbf{Chamfer-L1}:
\[
\mathrm{CD}_{L1} =
\frac{\mathrm{Acc}+\mathrm{Comp}}{2}.
\]

\noindent\textbf{Precision and Recall}:
For a distance threshold $\tau\in\{0.05,0.10\}$ m,
\[
\mathrm{Precision}_{\tau} =
\frac{1}{|\hat{P}|}
\sum_{\hat{\mathbf{p}}\in\hat{P}}
\mathbb{I}\left[
\min_{\mathbf{p}\in P}
\|\hat{\mathbf{p}}-\mathbf{p}\|_2 < \tau
\right],
\]
\[
\mathrm{Recall}_{\tau} =
\frac{1}{|P|}
\sum_{\mathbf{p}\in P}
\mathbb{I}\left[
\min_{\hat{\mathbf{p}}\in\hat{P}}
\|\mathbf{p}-\hat{\mathbf{p}}\|_2 < \tau
\right].
\]

\noindent\textbf{F-score}:
\[
\mathrm{F1}_{\tau} =
\frac{2\,\mathrm{Precision}_{\tau}\,
\mathrm{Recall}_{\tau}}
{\mathrm{Precision}_{\tau}+\mathrm{Recall}_{\tau}}.
\]
We report F1@5cm and F1@10cm using
$\tau=0.05$ m and $\tau=0.10$ m, respectively.

\begin{figure*}[!ht]
  \centering
  \includegraphics[width=0.8\linewidth]{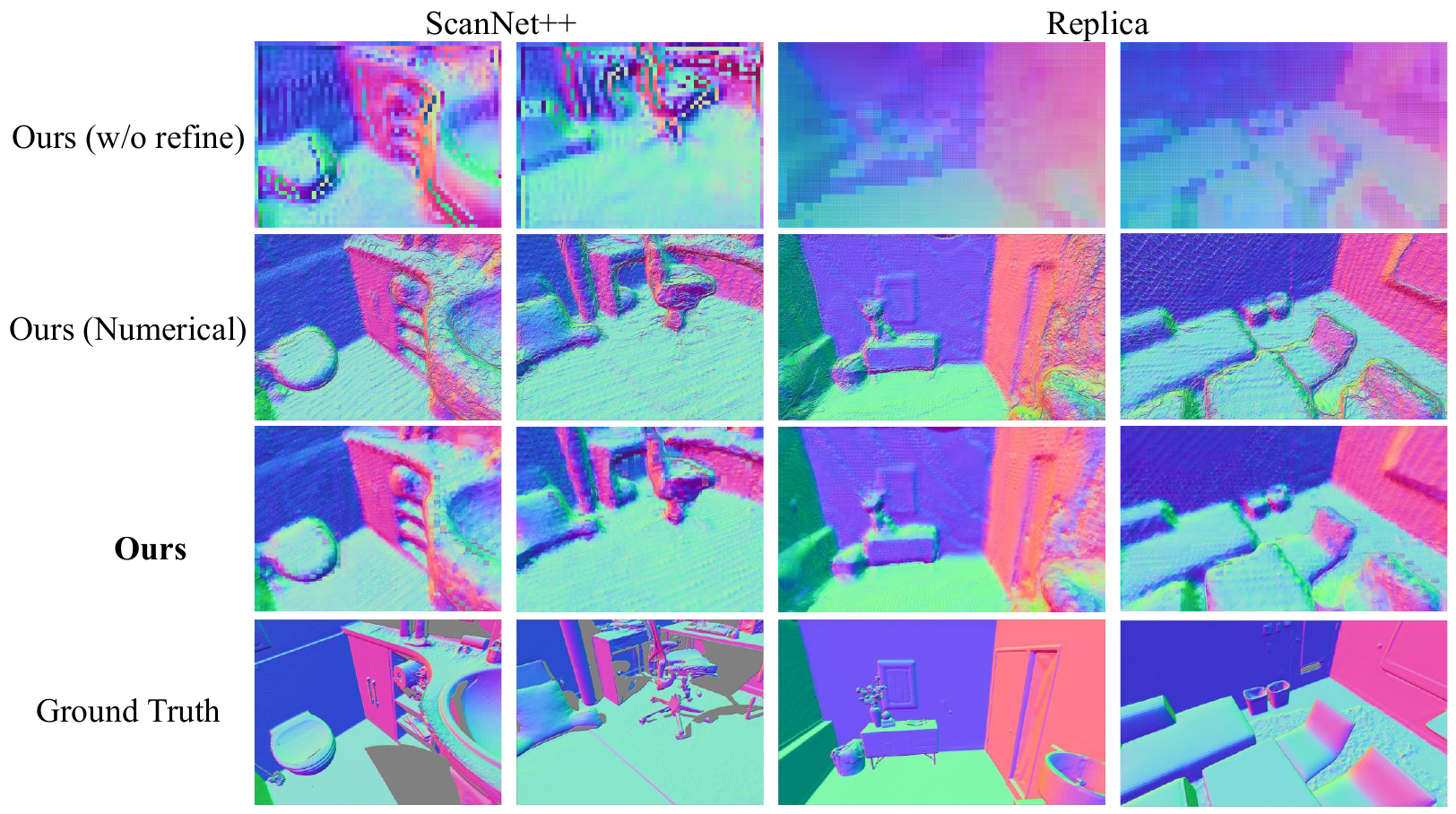}
  \centering
  \vspace{-0.2cm}
  \caption{Qualitative results of normal derivation on novel views by different techniques.}
  \label{fig:exp_qualitative_normal_num_ana_cmp}
\end{figure*}

\begin{figure*}[!ht]
  \centering
  \includegraphics[width=0.8\linewidth]{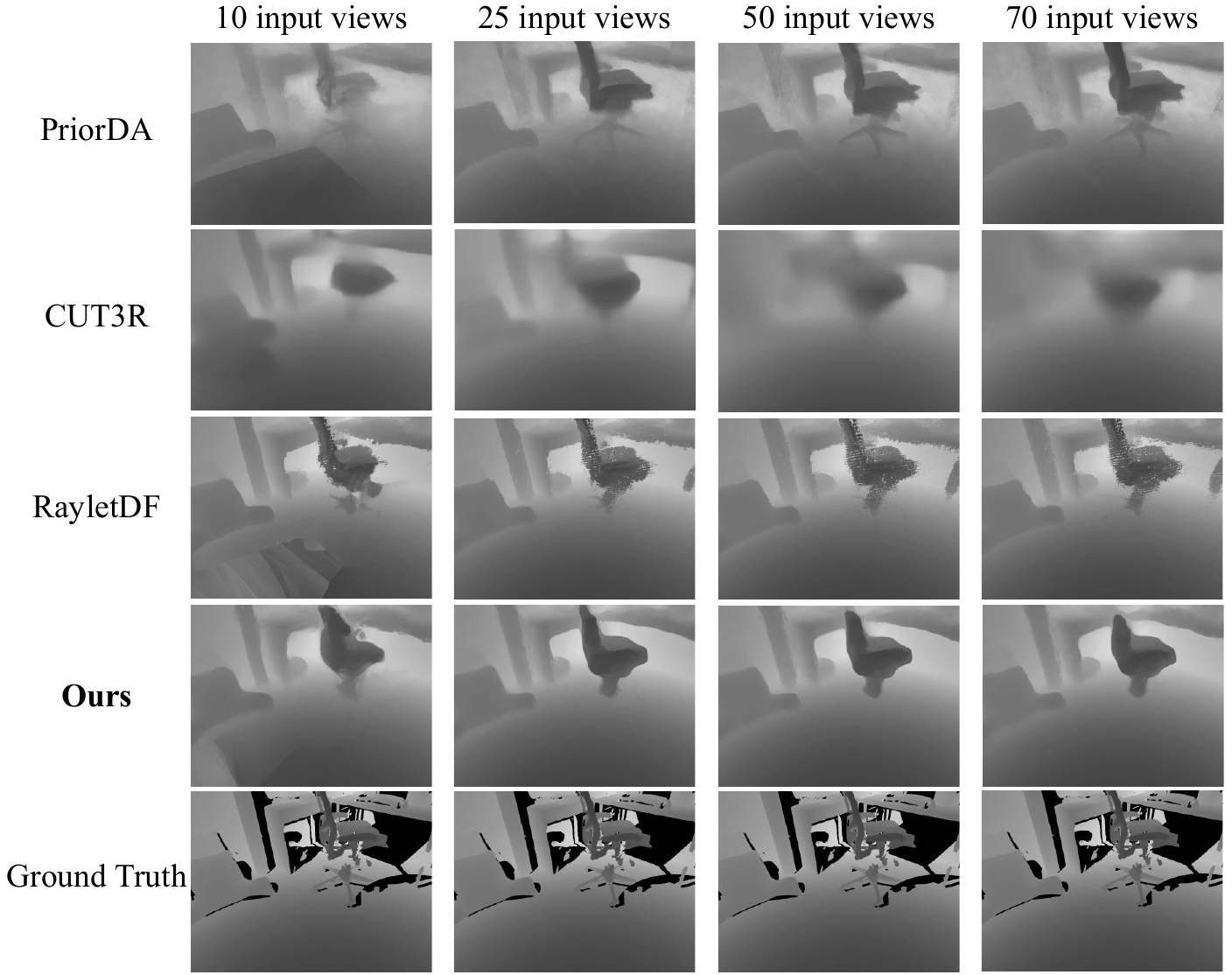}
  \centering
  \vspace{-0.2cm}
  \caption{Qualitative results of our method and baselines for 3D surface reconstruction under different input views.}
  \label{fig:exp_qualitative_app_abl_views}
\end{figure*}

\begin{figure*}[!ht]
  \centering
  \includegraphics[width=0.8\linewidth]{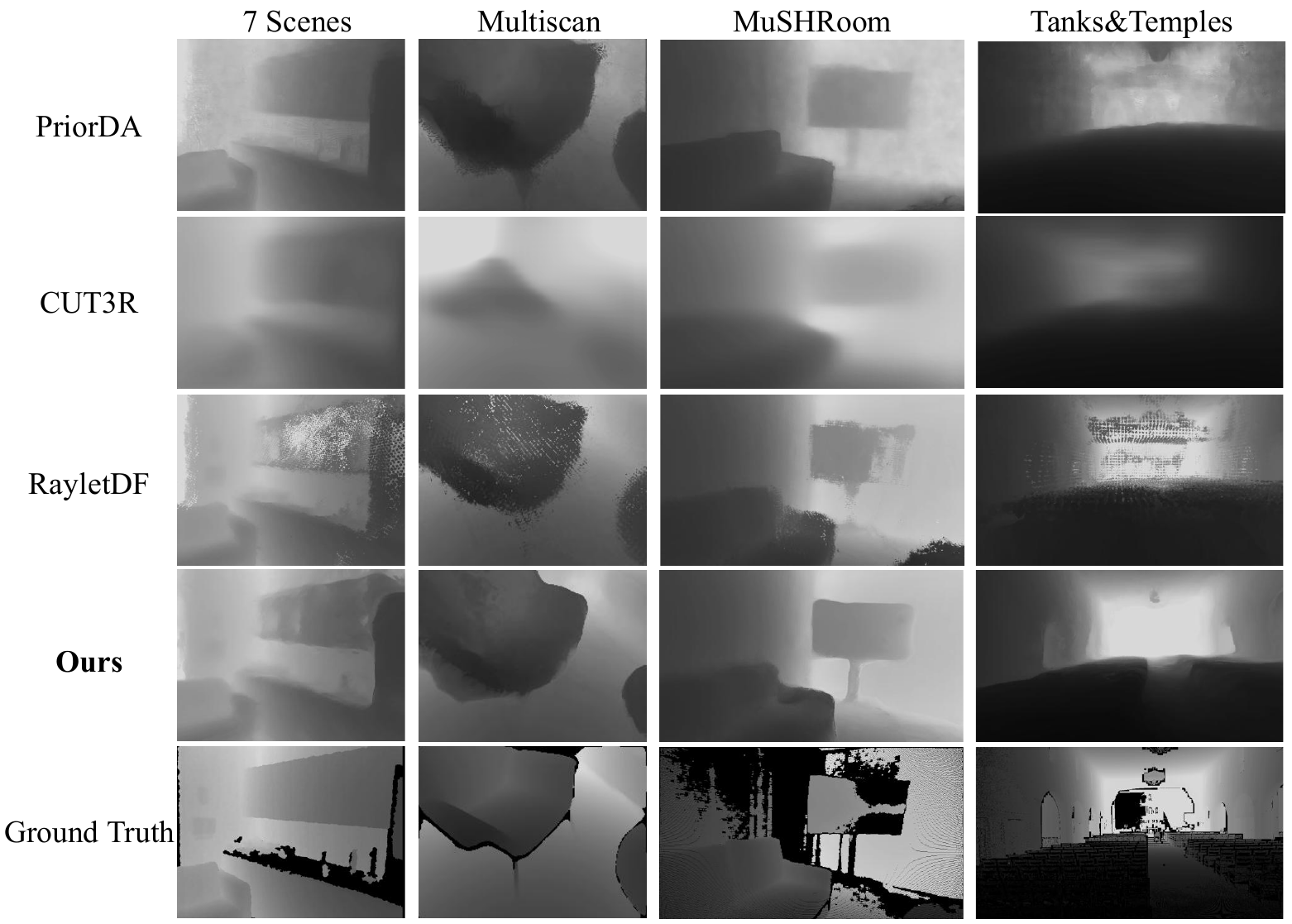}
  \centering
  \vspace{-0.2cm}
  \caption{Qualitative results of our method and baselines for 3D surface reconstruction on multiple datasets.}
  \label{fig:exp_qualitative_app_dist}
\end{figure*}

\begin{figure*}[!ht]
  \centering
  \includegraphics[width=0.8\linewidth]{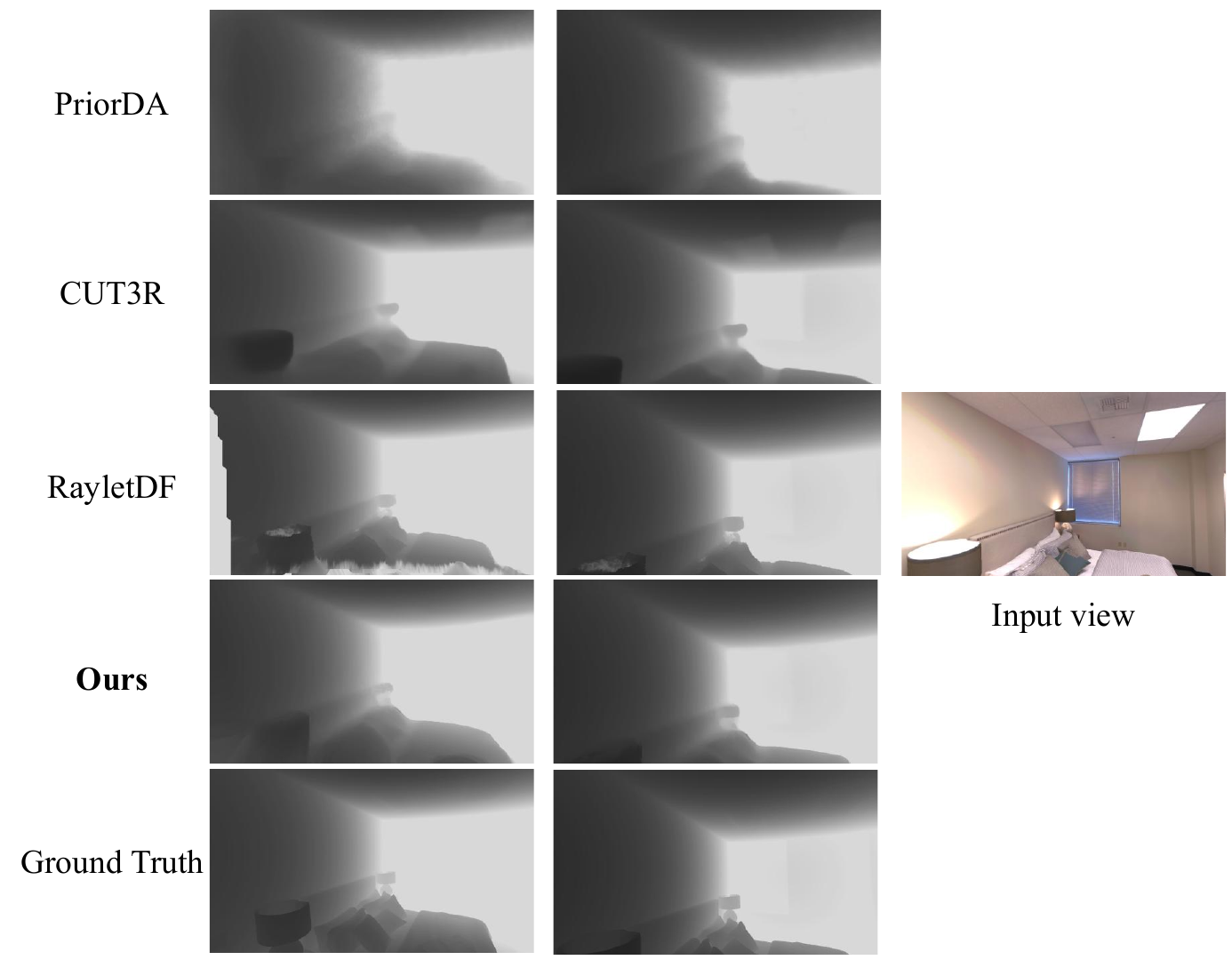}
  \centering
  \vspace{-0.2cm}
  \caption{Qualitative results of our method and baselines for single view 3D surface reconstruction on Replica.}
  \label{fig:exp_qualitative_app_single}
\end{figure*}

\subsection{Comparison of Normal Estimation}
\label{app:normal_cmp}
The qualitative results in Figure~\ref{fig:exp_qualitative_normal_num_ana_cmp} show a clear performance ranking. The coarse regressor produces a mosaic-like output lacking geometry details. Numerical differentiation of the refined field yields better but still noisy normals (with characteristic vein-like artifacts) compared to the clean analytical solution, highlighting the limitations of numerical methods.

\subsection{Qualitative Results of Reconstruction under Different Input Views}
\label{app:abl_views}
Qualitative results under varying numbers of input views are presented in Figure~\ref{fig:exp_qualitative_app_abl_views}. As shown, all methods except CUT3R improve as the number of views increases. Notably, \nickname{} exhibits a clear performance advantage over other baselines in sparse-view settings. In contrast, the predictions of CUT3R deteriorate significantly at 70 views, likely due to information loss caused by its compression mechanism.

\subsection{Qualitative Results of Single View Reconstruction}
\label{app:single_view}
The qualitative results for single-view reconstruction are shown in Figure~\ref{fig:exp_qualitative_app_single}. Notably, \nickname{} again demonstrates superior performance compared to other baselines, underscoring its robustness even under highly sparse input conditions.

\subsection{Additional Qualitative Results on Out-of-domain Datasts}
\label{app:add_vis}
We show more qualitative results of out-of-domain datasets in Figure~\ref{fig:exp_qualitative_app_dist}.

\begin{figure*}[ht]
  \centering
  \includegraphics[width=1\linewidth]{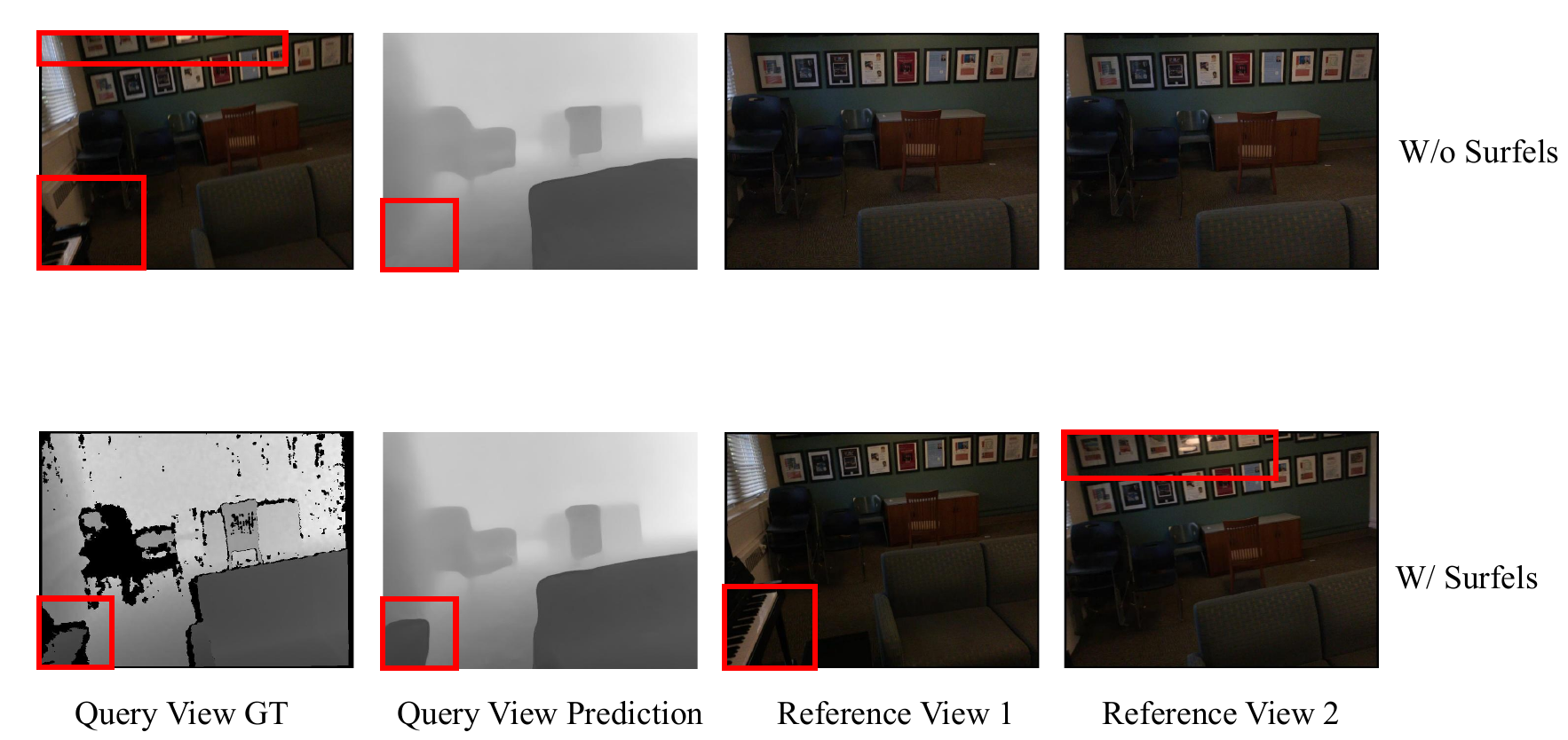}
  \centering
  \vspace{-0.2cm}
  \caption{Qualitative comparison of surfel-based reference selection.}
  \label{fig:exp_qualitative_app_surfel}
\end{figure*}

\subsection{Qualitative Results on Surfel Voting}
\label{app:surfel_abl}
A case study in Figure~\ref{fig:exp_qualitative_app_surfel} illustrates a limitation of per-pixel comparison for view selection (first row): it fails to capture some corner geometry (highlighted in red). In contrast, our surfel-based voting approach (second row) effectively covers these regions, validating its robustness in capturing widespread geometry.

\subsection{Additional Quantitative Ablation Experiment Results}
\label{app:scannet_abl}
We provide ablation results on ScanNet dataset in Table~\ref{tab:app_exp_abl} and Table~\ref{tab:app_exp_abl_topk}. The performance gains from full pipeline ablations on in-domain ScanNet are less pronounced than those on the challenging MuSHRoom dataset. Nevertheless, the results still demonstrate that models incorporating our raylet-based surface module consistently achieve superior performance across most configurations. These findings validate the robust representational capacity of the raylet-based architecture. Notably, the significant sensitivity of the coarse ray-surface regressor's accuracy to these components underscores their critical role in ensuring efficient and precise estimation. As shown in Table~\ref{tab:app_exp_abl_topk}, on in-domain dataset ScanNet $\textit{top-1}$ or $\textit{top-2}$ is sufficient. We also investigate the $\textit{top-k}$ robustness of our model trained with 2 reference views on MuSHRoom dataset, results are summarized in Table~\ref{tab:exp_abl_robust_topk}. As shown, a large number of reference views is not required.

\begin{table}[h] 
\tabcolsep=0.15cm  
\centering
\caption{Ablation results for evaluating on the test split of ScanNet.}
\label{tab:app_exp_abl}
\resizebox{0.8\linewidth}{!}
{
\begin{tabular}{llccccc}
\toprule[1.0pt]
&& ADE$\downarrow$ &RMSE$\downarrow$ &Abs-Rel$\downarrow$&Sq-Rel$\downarrow$&$\delta\uparrow $  \\
\toprule[1.0pt]

\multirow{2}{*}{Full}&w/o Refine &0.084&0.169&0.044&\textbf{0.019}&\textbf{0.969}\\
&w/ Refine  &\textbf{0.081}&0.171&\textbf{0.042}&0.020&\textbf{0.969}\\
\midrule
\multirow{2}{*}{w/o surfel vote}&w/o Refine &0.086&0.171&0.045&0.020&0.968\\
&w/ Refine  &\textbf{0.081}&0.173&0.043&0.020&0.968\\
\midrule
\multirow{2}{*}{w/o local coordinate}&w/o Refine &0.120&0.213&0.066&0.031&0.945\\
&w/ Refine  &0.098&0.202&0.053&0.029&0.953\\
\midrule
\multirow{2}{*}{w/o hierarchy feature}&w/o Refine &0.090&0.176&0.047&0.021&0.967\\
&w/ Refine  &0.082&0.175&0.043&0.021&0.968\\
\midrule
\multirow{2}{*}{w/o \textit{top-k} reference view}&w/o Refine &0.094&0.180&0.049&0.021&0.966\\
&w/ Refine  &\textbf{0.081}&\textbf{0.170}&\textbf{0.042}&\textbf{0.019}&\textbf{0.969}\\
\toprule[1.0pt]
\end{tabular}
}
\end{table}

\begin{table}[h] 
\tabcolsep=0.15cm  
\centering
\caption{\textit{Top-k} ablation results for evaluating on the test split of ScanNet.}
\label{tab:app_exp_abl_topk}
\resizebox{0.8\linewidth}{!}
{
\begin{tabular}{llccccc}
\toprule[1.0pt]
&& ADE$\downarrow$ &RMSE$\downarrow$ &Abs-Rel$\downarrow$&Sq-Rel$\downarrow$&$\delta\uparrow $  \\
\toprule[1.0pt]

\multirow{2}{*}{\textit{top-1}}&w/o Refine &0.082&0.169&0.043&\textbf{0.019}&\textbf{0.969}\\
&w/ Refine  &\textbf{0.080}&0.171&0.043&0.020&0.968\\
\midrule
\multirow{2}{*}{\textit{top-2}}&w/o Refine&0.084&0.169&0.044&0.020&\textbf{0.969}\\
&w/ Refine  &0.081&0.173&0.043&0.020&0.968\\
\midrule
\multirow{2}{*}{w/o \textit{top-k} reference view}&w/o Refine &0.094&0.180&0.049&0.021&0.966\\
&w/ Refine  &0.081&\textbf{0.170}&\textbf{0.042}&\textbf{0.019}&\textbf{0.969}\\
\toprule[1.0pt]
\end{tabular}
}
\end{table}

\vspace{-0.5cm}
\begin{table}[!h] 
\tabcolsep=0.15cm  
\centering
\captionsetup{skip=2pt}
\caption{Robustness of \textit{top-k} on the test split of MuSHRoom.}
\label{tab:exp_abl_robust_topk}
\resizebox{0.6\linewidth}{!}
{
\begin{tabular}{lccccc}
\toprule[1.0pt]
& ADE$\downarrow$ &RMSE$\downarrow$ &Abs-Rel$\downarrow$&Sq-Rel$\downarrow$&$\delta\uparrow $  \\
\toprule[1.0pt]
\textit{top-1}     &0.341&0.548&0.272&\textbf{0.371}&0.756\\
\textit{top-2}     &\textbf{0.331}&\textbf{0.538}&\textbf{0.266}&0.372&\textbf{0.761}\\
\textit{top-5}   &0.337 &0.540&0.270&0.367 &0.753\\
\textit{top-10}   &0.352 &0.555&0.287&0.393 &0.738\\
\toprule[1.0pt]
\end{tabular}
}\vspace{-0.5cm}
\end{table}

\subsection{Evaluation of In-domain Reconstruction}
\label{app:indomain_eval}
In this setting, we further evaluate the performance of \nickname{} on in-domain dataset. ScanNet is selected as evaluation dataest because it is within all of CUT3R, VGGT and our training dataset. We also uniformly sample 50 images as input which is the same as the out-of-domain reconstruction evaluation.
The quantitative results are reported in Table~\ref{tab:exp_test_scan}, our \nickname{} demonstrates clearly better reconstruction accuracy on the in-domain dataset, which is similar to Section~\ref{sec:outdomain_eval}.

\begin{table}[th] 
\tabcolsep=0.15cm  
\centering
\caption{Qualitative results for in-domain dataset evaluating on the test split of ScanNet.}
\label{tab:exp_test_scan}
\resizebox{0.6\linewidth}{!}
{
\begin{tabular}{lccccc}
\toprule[1.0pt]
& ADE$\downarrow$ &RMSE$\downarrow$ &Abs-Rel$\downarrow$&Sq-Rel$\downarrow$&$\delta\uparrow $  \\
\toprule[1.0pt]

Projection~\cite{Wang2025a} &0.247&0.504&0.124&0.139&0.876\\
PriorDA~\cite{wang2025depthprior}  &0.125&0.240&0.057&0.029&0.937\\
CUT3R~\cite{Wang2025}& 0.298&0.411&0.167&0.114&0.780\\
FLARE~\cite{Zhang2025d} &0.238 &0.435 &0.114 &0.115 &0.855\\
RayletDF~\cite{Wei2025} &0.136&0.271&0.071&0.059&0.919\\
Ours \textit{w/o refine} &0.105&0.195&0.055&\textbf{0.025}&0.961\\
\textbf{Ours} &\textbf{0.089}&\textbf{0.188}&\textbf{0.047}&\textbf{0.025}&\textbf{0.962}\\
\toprule[1.0pt]
\end{tabular}
}
\end{table}

\subsection{Details of Datasets}
\label{app_dataset}
\subsubsection{ScanNet++}
We adopt the official test splits for novel view synthesis but exclude scene “1ada7a0617” due to corruptions in its COLMAP data . For each scene, we uniformly sample every 5th frame from the video sequence for training, and reserve every 10th frame in the test set for testing. We use the iPhone sequences with COLMAP registered poses and render mesh scans to obtain high-resolution depth maps. All the images are resized to 960$\times$720.

\subsubsection{ScanNet}
We use the official training and test splits of ScanNetV2 dataset, with 1201 training scenes and 100 test scenes. Similar to ScanNet++, we also load every 5th frame from the trajectories for training, and then reserve every 10th frame in test set for evaluation. 

\subsubsection{ARKitScenes}
We use the training split of DUSt3R. For training views, we sample every 10th RGB image from the ``vga-wide" dataset set and unsample the depth to 640$\times$480 as in DUSt3R.

\subsubsection{BlendMVS}
All the images of 507 scenes in raw dataset are used in our training dataset. For convenience of sampling during training, we calculate the overlap between each image pair and employ greedy algorithm to rank images in every scenes in video frame sequence order.

\subsubsection{HyperSim}
We use the ``cam\_00" trajectory in each scene and all the images are used for training. Similar to BlendMVS images are ranked in video frame sequence order for training convenience.
\subsubsection{WildRGBD}
Given the extensive scale of the dataset, we select the first ten scenes from each category to construct our training set. Following the preprocessing pipeline in DUSt3R, all images are cropped and resized to a fixed 4:3 aspect ratio to maintain consistency across samples.
\subsubsection{MVS-Synth}

The official raw dataset, featuring a resolution of 960$\times$540, is employed. All images within the sequences are incorporated into our training framework without further downsampling.
\subsubsection{S3DIS}
Our training set comprises all areas except for Area 5 and every 10th frame is reserved for validation. Since the raw data is unordered, we employ a custom ranking algorithm to establish a coherent sequence for training. 
\subsubsection{SceneNN}
All scenes in the dataset are utilized for training. To reduce temporal redundancy while maintaining scene coverage, we sample frames from each trajectory at a stride of 10.
\subsubsection{GTASfM}
We adhere to the official training split of GTASfM. All images within the designated split are used to construct our training dataset.
\subsubsection{Multiscan}
We use official test split of Multiscan, loading every 50th image from trajectory and then reserve every 10th frame in test set for evaluation.
\subsubsection{Replica}
We use the pre-processed Replica dataset provided by DN-Splatter~\cite{Turkulainen2025}, which contains 8 scenes with 1200$\times$680 resolution images. Each scene consists of 64 training images and 16 testing images.
\subsubsection{7Scenes}
Since the original ground truth is imperfect, the same as~\cite{wang2025amb3r}, we adopt COLMAP GT~\cite{Brachmann2021ICCV} provided that is obtained via global optimization across all sequences.For evaluation, we use the 18-scene split, consistent with the setup in Spann3R~\cite{wang20243d}.
\subsubsection{MuSHRoom}
We employ the official MuSHRoom split, using the long-sequence RGB-D data captured by Kinect sensors. Every 10th frame is sampled for testing.
\subsubsection{Tanks\&Temples}
Data is sourced from the RMVDB~\cite{schroeppel2022robust} variant, which provides high-quality processed depth maps. For evaluation purposes, we reserve every 10th frame of the sequences as the test set.

\subsection{Details of Sparse Convolution and Raylet Distance Network}
\label{app_spconv}
\begin{figure}[!ht]
  \centering
  \includegraphics[width=0.3\linewidth]{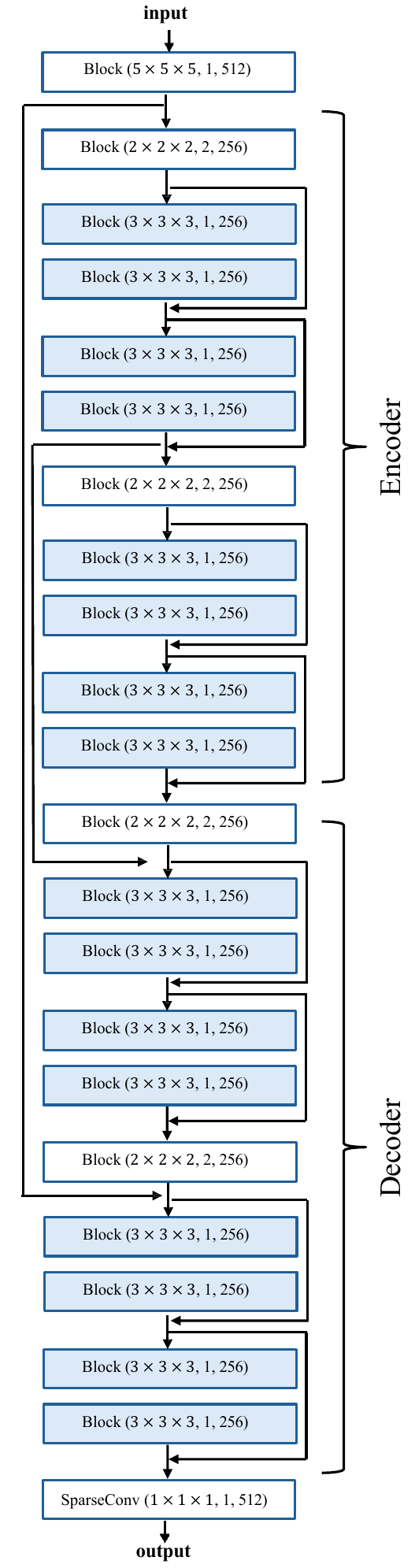}
  \centering
  \vspace{-0.2cm}
  \caption{Details of U-Net feature  extractor. Block means a sequential application of SparseConvolution in encoder or  SparseInverseConvolution in decoder, GroupNorm and ReLU activation.}
  \label{fig:spconv}
\end{figure}

We simply employ the SparseConv architecture as our Feature extractor. Particularly, we use the existing U-Net implementation of SPconv PyTorch package \url{https://github.com/traveller59/spconv}. As illustrated in Figure~\ref{fig:spconv}, the encoder and decoder have 8 layers respectively, which produce 512-dimensional features. Block means a sequential application of SparseConvolution in encoder or  SparseInverseConvolution in decoder, GroupNorm and ReLU activation.

Our Raylet distance network architecture consists of: 1) an initial MLP layer with ReLU that projects input features to 512 dimensions; 2) a block of 8 successive dense layers (512 neurons, ReLU); and 3) a final linear output layer that predicts the raylet distance.

\subsection{Robustness Evaluation of the S-nearest Neighbor}
To evaluate the robustness of our S-nearest neighbor search against potential outliers, we progressively downsample the pointmaps generated by VGGT to simulate varying levels of data degradation. Specifically, we retain only a certain percentage of the pointmaps (e.g., 80\%, 40\%) as input to our method. As shown in the following Table \ref{tab:knn}, our method demonstrates strong robustness against these challenging conditions, essentially because the voxelization process is particularly robust to potential outliers. 

\vspace{-0.5cm}
\begin{table}[!h] 
\tabcolsep=0.15cm  
\centering
\caption{Quantitative results under varying levels of downsampled data prior on ScanNet++.}
\label{tab:knn}

\begin{tabular}{lccccc}
\toprule[1.0pt]
& ADE$\downarrow$ &RMSE$\downarrow$ &Abs-Rel$\downarrow$&Sq-Rel$\downarrow$&$\delta\uparrow $  \\
\toprule[1.0pt]
100\%     &0.132&0.259&0.079&0.067&0.910\\
80\%     &0.132&0.259&0.079&0.067&0.910\\
60\%     &0.132&0.259&0.079&0.067&0.910\\
40\%     &0.132&0.259&0.079&0.067&0.910\\
20\%     &0.132&0.259&0.079&0.067&0.910\\
5\%     &0.133&0.260&0.080&0.068&0.909\\
1\%     &0.134&0.261&0.080&0.069&0.907\\
\toprule[1.0pt]
\end{tabular}

\end{table}

\subsection{ Qualitative Results of 3D Outputs}
\begin{figure}[t]
  \centering
  \includegraphics[width=0.9\linewidth]{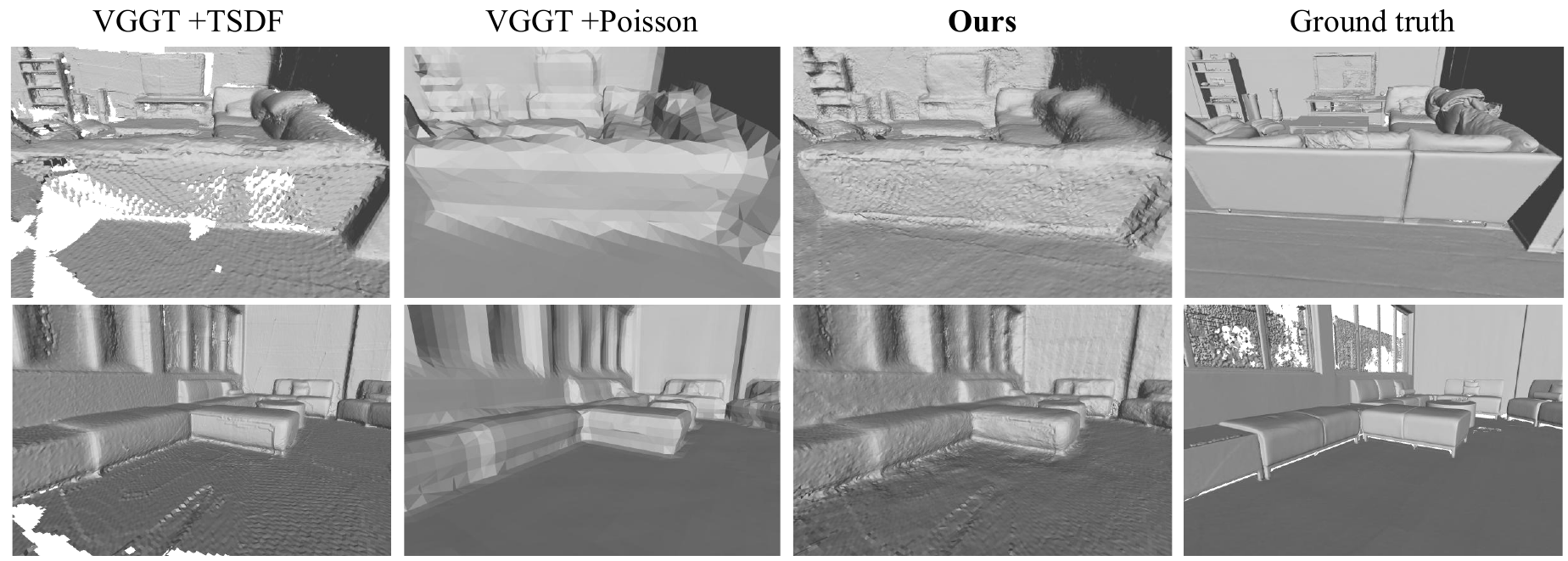}
    \includegraphics[width=0.9\linewidth]{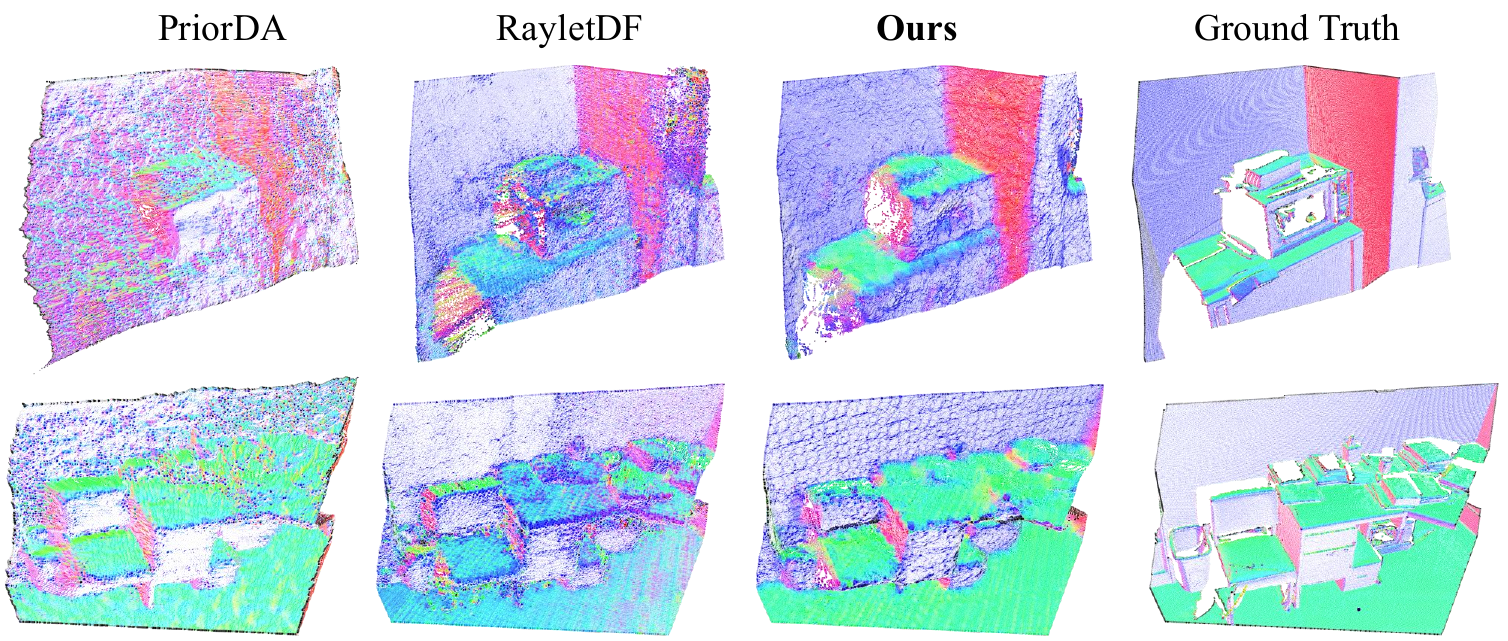}
  \centering
  \caption{Qualitative 3D results of our method and baselines on ScanNet++. The upper two rows compare 3D meshes, and the bottom two rows compare 3D point clouds.}
  \label{fig:3d_vis}
\end{figure}

we provide qualitative results of the reconstructed 3D meshes and point clouds in the attached Figure~\ref{fig:3d_vis}.

\subsection{ Failure Cases}
\begin{figure}[ht]
  \centering
  \includegraphics[width=0.9\linewidth]{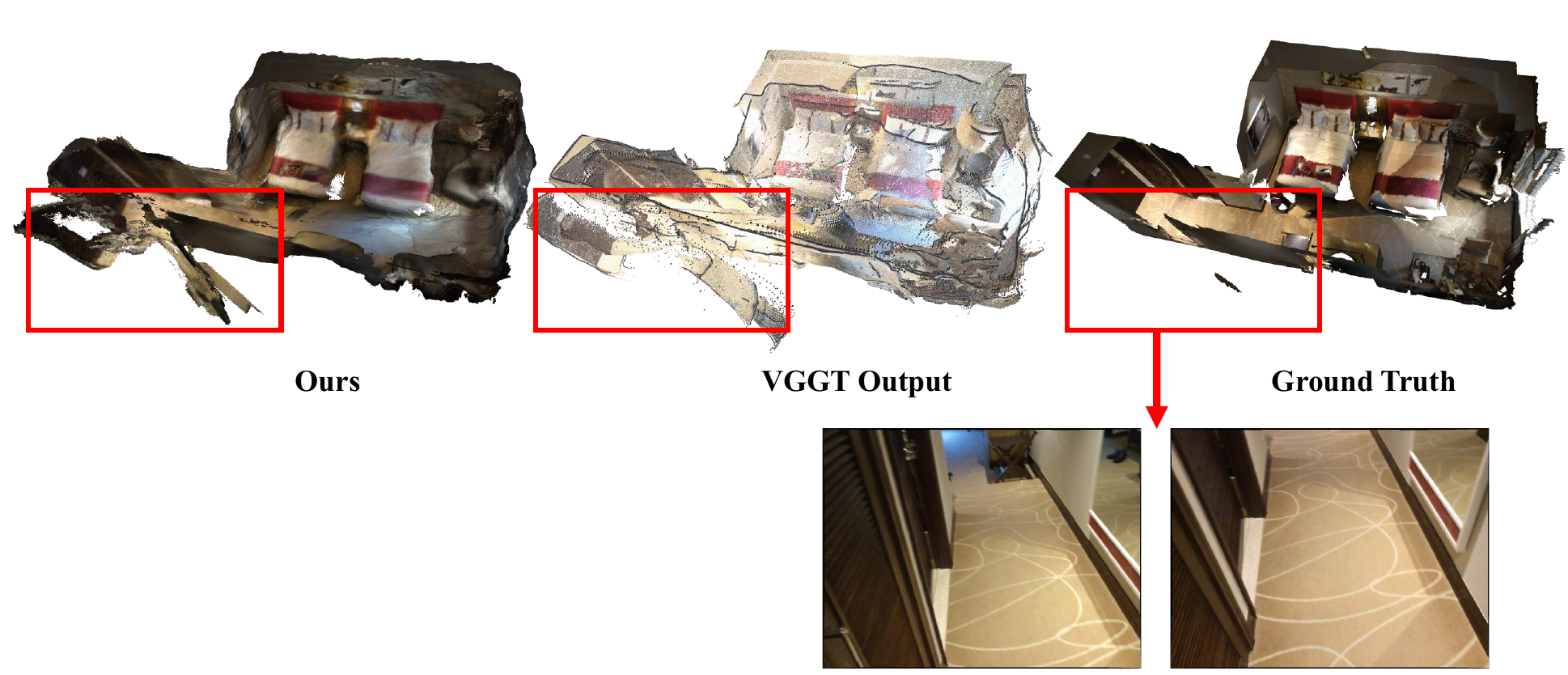}
    \vspace{0.3cm}
  \includegraphics[width=0.8\linewidth]{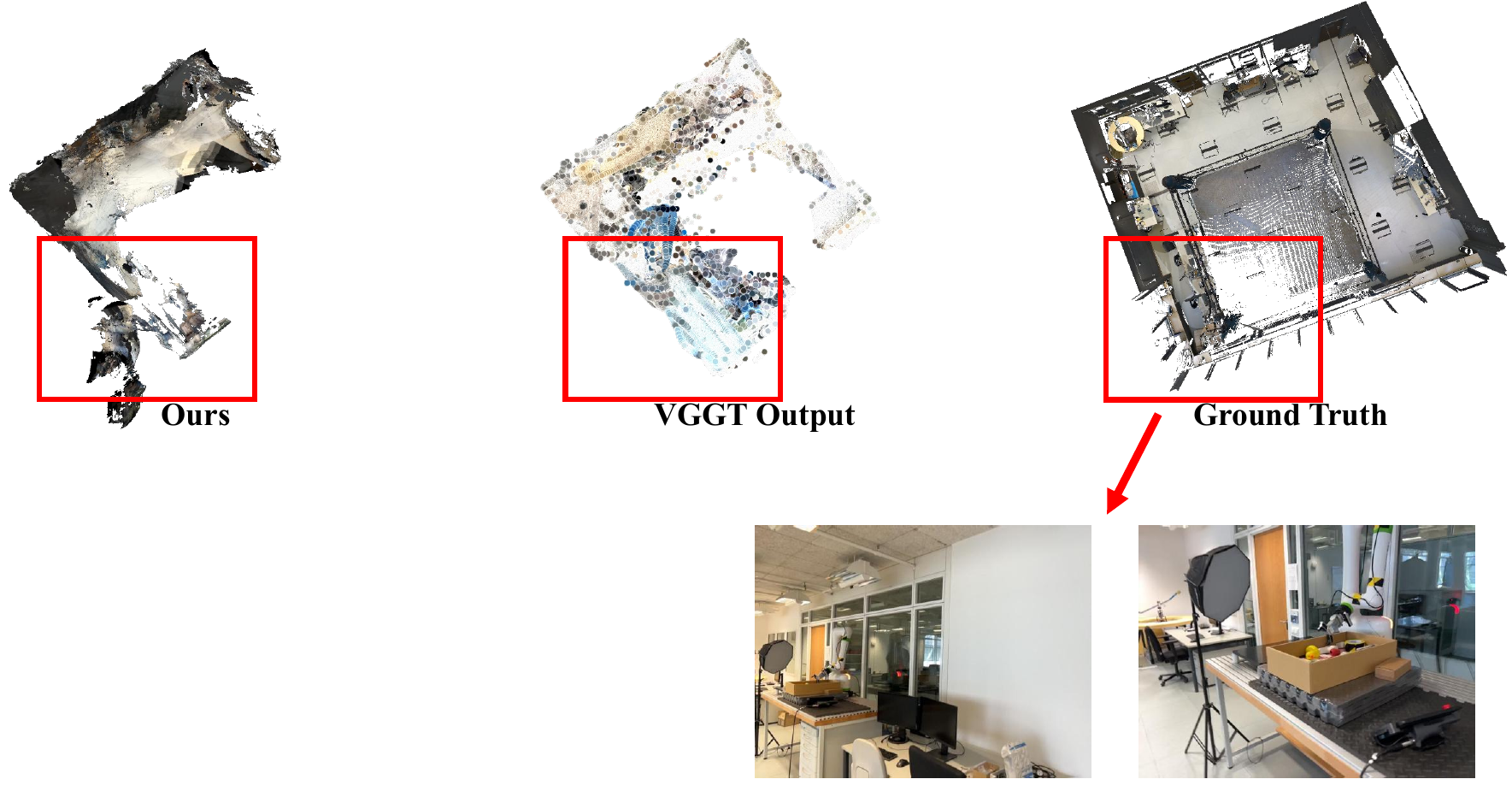}
  \centering
  \caption{Failure cases of our method. For example, the mirror and transparent glasses on the wall cause inaccurate pointmap estimation by VGGT, which ultimately leads to errors in our approach.}
  \label{fig:fail_case}
\end{figure}

Additionally, we include failure cases in the attached Figure~\ref{fig:fail_case}. For example, the \textit{mirror} and \textit{transparent glasses} on the wall cause VGGT to generate incorrect pointmaps, ultimately leading to the failure of our method.

\subsection{Effectiveness of Raylet-Based Refinement}
To quantitatively assess the contribution of our raylet-based refinement module, particularly in high-frequency regions, we applied the Canny edge detector to the RGB images (test views) to categorize all pixels into two groups: 1) edge pixels, representing high-frequency regions, and 2) non-edge pixels, representing low-frequency regions. The following Table \ref{tab:high_low} reports the performance with and without the raylet-based refinement module across both high- and low-frequency regions on Replica dataset. We can see that our refinement module can consistently improve the surface reconstruction accuracy in both high- and low-frequency regions, demonstrating its general effectiveness.

\begin{table}[!h] 
\tabcolsep=0.15cm  
\centering
\caption{Quantitative results of high/low frequency regions on Replica.}
\label{tab:high_low}

\begin{tabular}{l|lccccc}
\toprule[1.0pt]
& & ADE$\downarrow$ &RMSE$\downarrow$ &Abs-Rel$\downarrow$&Sq-Rel$\downarrow$&$\delta\uparrow $  \\
\toprule[1.0pt]
high freq & w/o refine     &0.173&0.283&0.068&0.038&0.949\\
& w/ refine     &0.116&0.258&0.048&0.037&0.951\\
\midrule
low freq & w/o refine      &0.132&0.202&0.051&0.018&0.980\\
& w/ refine     &0.057&0.145&0.024&0.012&0.985\\
\toprule[1.0pt]
\end{tabular}

\end{table}

\begin{table}[!h] 
\tabcolsep=0.15cm  
\centering
\caption{The memory consumption and rendering speed given different numbers of input views.}
\label{tab:mem_view}
\resizebox{0.6\linewidth}{!}
{
\begin{tabular}{ccc}
\toprule[1.0pt]
\# of input views& Render Speed (FPS) & GPU Memory (GB)  \\
\toprule[1.0pt]
25    &9.29 & 16.52 \\
50    &8.68 & 20.89 \\
70  &8.19 & 23.60\\
100     &7.68& 28.06\\
\toprule[1.0pt]
\end{tabular}
}
\end{table}

\subsection{Efficiency of Ray-Based Surface Queries}
\label{app:target_view_efficiency}

Unlike TSDF, Poisson, and NKSR, which require explicit surface reconstruction before querying target views, our feed-forward ray-based model directly predicts ray-surface distances without explicit mesh construction. As shown in Table~\ref{tab:rebuttal_table10_time}, our method achieves the lowest target-view query time (0.179 s), demonstrating the efficiency of our direct surface querying approach.

\begin{table}[!h]
\centering
\caption{Time required to obtain a target view from the VGGT output.}
\label{tab:rebuttal_table10_time}
\setlength{\tabcolsep}{0.25cm}
\begin{tabular}{lc}
\toprule
Method & Time (seconds)$\downarrow$ \\
\midrule
VGGT + TSDF~\cite{curless1996volumetric} & 0.216 \\
VGGT + Poisson~\cite{kazhdan2013screened} & 55.564 \\
VGGT + NKSR~\cite{huang2023neural} & 3.924 \\
\textbf{Ours} & \textbf{0.179} \\
\bottomrule
\end{tabular}
\end{table}

\subsection{Additional Comparison with Mesh Reconstruction Methods}

We further compare our method with conventional mesh reconstruction approaches, including TSDF, Poisson, and NKSR, across six datasets. Specifically, we reconstruct explicit meshes from VGGT-predicted pointmaps and render depth maps at the query views to obtain ray--surface distances. All methods are evaluated using the same metrics as in the main paper. The quantitative results are reported in Table~\ref{app:exp_ray_surface_dis_mesh}.

\begin{table}[h]\tabcolsep=0.1cm 
\centering
\captionsetup{skip=2pt}
\caption{Quantitative results of estimated ray-surface distances using meshes reconstructed from VGGT pointmaps.}
\resizebox{1\textwidth}{!}{
\begin{tabular}{@{}r|ccccc|ccccc|ccccc@{}}
    \toprule
   &\multicolumn{5}{c|}{\textit{test on} $\rightarrow$ ScanNet++}&\multicolumn{5}{c|}{\makecell{\textit{test on} $\rightarrow$ 7Scenes}}&\multicolumn{5}{c}{\makecell{\textit{test on} $\rightarrow$ MultiScan}}\\
   \cmidrule{2-16}
   &ADE$\downarrow$ &RMSE$\downarrow$ &Abs-Rel$\downarrow$&Sq-Rel$\downarrow$&$\delta\uparrow $&ADE$\downarrow$&RMSE$\downarrow$&Abs-Rel$\downarrow$&Sq-Rel$\downarrow$&$\delta\uparrow$&
   ADE$\downarrow$&RMSE$\downarrow$&Abs-Rel$\downarrow$&Sq-Rel$\downarrow$&$\delta\uparrow$ \\
   \midrule
    VGGT + TSDF~\cite{curless1996volumetric}&
   0.436&0.777&0.235&0.404&0.766
   &0.228&0.489&0.113&0.135&0.885   &0.794&0.986&0.620&0.803&0.381\\
   VGGT + Poisson~\cite{kazhdan2013screened}&
   0.175&0.327&0.094&0.106&0.881& 0.165&0.307&0.081&0.048&0.888&0.384&0.482&0.296&0.276&0.616 \\
    VGGT + NKSR~\cite{huang2023neural}&
   0.446&0.648&0.231&0.298&0.656 &0.213&0.371&0.104&0.072&0.832    &0.509&0.626&0.379&0.379&0.489\\

    Ours \textit{w/o refine}&
   0.162&0.270&0.097&0.063&0.894 &0.130&\textbf{0.230}&0.068&\textbf{0.029}&\textbf{0.945}    &0.233&0.309&0.222&0.147&0.737\\
   \textbf{Ours} &
   \textbf{0.132}&\textbf{0.259}&\textbf{0.079}&\textbf{0.067}&\textbf{0.910}& \textbf{0.123}&0.233&\textbf{0.064}&0.031&0.943 &\textbf{0.205}&\textbf{0.287}&\textbf{0.190}&\textbf{0.137}&\textbf{0.774}\\
   \midrule
    &\multicolumn{5}{c|}{\textit{test on} $\rightarrow$ Replica}&\multicolumn{5}{c|}{\makecell{\textit{test on} $\rightarrow$ MuSHRoom}}&\multicolumn{5}{c}{\makecell{\textit{test on} $\rightarrow$ Tanks\&Temples}}\\
   \cmidrule{2-16}
   VGGT + TSDF~\cite{curless1996volumetric}&
   0.116&0.366&0.050&0.074&0.958& 0.606&1.009&0.348&0.652&0.657&1.391&3.577&\textbf{0.112}&\textbf{1.741}&\textbf{0.910} \\
   VGGT + Poisson~\cite{kazhdan2013screened}&
   \textbf{0.063}&0.182&\textbf{0.024}&\textbf{0.015}&0.979& 0.284&0.509&\textbf{0.170}&\textbf{0.233}&0.795&2.816&4.123&0.186&2.216&0.757 \\
    VGGT + NKSR~\cite{huang2023neural}&
   0.150&0.348&0.052&0.051&0.914 &0.908&1.187&0.407&0.709&0.453    &5.701&6.879&0.441&4.386&0.379\\

    Ours \textit{w/o refine}&
   0.137&0.215&0.053&0.021&0.976 &0.350&0.511&0.299&0.430&0.734  &1.622&3.144&0.183&1.929&0.865\\
   \textbf{Ours}  & 
   0.064&\textbf{0.165}&0.027&\textbf{0.015}&\textbf{0.981} &\textbf{0.254}&\textbf{0.458}&0.232&0.429&\textbf{0.822 }&\textbf{1.233}&\textbf{2.977}&\textbf{{0.144}}&2.012&0.908\\
\bottomrule
\end{tabular}}
\label{app:exp_ray_surface_dis_mesh} \vspace{-0.5cm}
\end{table}


\clearpage
\section*{NeurIPS Paper Checklist}

\begin{enumerate}

\item {\bf Claims}
    \item[] Question: Do the main claims made in the abstract and introduction accurately reflect the paper's contributions and scope?
    \item[] Answer: \answerYes{} 
    \item[] Justification: The main contribution of our paper has been clearly declared in the abstract and introduction. 

\item {\bf Limitations}
    \item[] Question: Does the paper discuss the limitations of the work performed by the authors?
    \item[] Answer: \answerYes{} 
    \item[] Justification: We include a limitation section. 

\item {\bf Theory assumptions and proofs}
    \item[] Question: For each theoretical result, does the paper provide the full set of assumptions and a complete (and correct) proof?
    \item[] Answer: \answerNA{} 
    \item[] Justification: We are proposing a comprehensive framework, not including theoretical results. 

    \item {\bf Experimental result reproducibility}
    \item[] Question: Does the paper fully disclose all the information needed to reproduce the main experimental results of the paper to the extent that it affects the main claims and/or conclusions of the paper (regardless of whether the code and data are provided or not)?
    \item[] Answer: \answerYes{} 
    \item[] Justification: A detailed description of our framework is provided, and the code will be released upon acceptance.

\item {\bf Open access to data and code}
    \item[] Question: Does the paper provide open access to the data and code, with sufficient instructions to faithfully reproduce the main experimental results, as described in supplemental material?
    \item[] Answer: \answerYes{} 

\item {\bf Experimental setting/details}
    \item[] Question: Does the paper specify all the training and test details (e.g., data splits, hyperparameters, how they were chosen, type of optimizer) necessary to understand the results?
    \item[] Answer: \answerYes{} 
    \item[] Justification: We discuss all the experiment details both in the Experiment section and in the Appendix.

\item {\bf Experiment statistical significance}
    \item[] Question: Does the paper report error bars suitably and correctly defined or other appropriate information about the statistical significance of the experiments?
    \item[] Answer: \answerYes{} 
    \item[] Justification: We take the experiments with random initialization on several different cases and take the average. 

\item {\bf Experiments compute resources}
    \item[] Question: For each experiment, does the paper provide sufficient information on the computer resources (type of compute workers, memory, time of execution) needed to reproduce the experiments?
    \item[] Answer: \answerYes{} 
    \item[] Justification: We provide sufficient information on the computer resources in the Appendix. 
    
\item {\bf Code of ethics}
    \item[] Question: Does the research conducted in the paper conform, in every respect, with the NeurIPS Code of Ethics \url{https://neurips.cc/public/EthicsGuidelines}?
    \item[] Answer: \answerYes{} 
    \item[] Justification: We follow the NeurIPS Code of Ethics.

\item {\bf Broader impacts}
    \item[] Question: Does the paper discuss both potential positive societal impacts and negative societal impacts of the work performed?
    \item[] Answer: \answerYes{} 
    \item[] Justification: We discussed the potential positive impact of our work, as we outlined some of its application in the introduction section of our main text. As for negative social impacts, we have not yet identified significant negative implications.
    
\item {\bf Safeguards}
    \item[] Question: Does the paper describe safeguards that have been put in place for responsible release of data or models that have a high risk for misuse (e.g., pre-trained language models, image generators, or scraped datasets)?
    \item[] Answer: \answerNA{} 
    \item[] Justification: This paper poses no such risks. 

\item {\bf Licenses for existing assets}
    \item[] Question: Are the creators or original owners of assets (e.g., code, data, models), used in the paper, properly credited and are the license and terms of use explicitly mentioned and properly respected?
    \item[] Answer: \answerYes{} 
    \item[] Justification: We have properly cited all the assets we used in the paper. 

\item {\bf New assets}
    \item[] Question: Are new assets introduced in the paper well documented and is the documentation provided alongside the assets?
    \item[] Answer: \answerNA{} 
    \item[] Justification: This paper does not release new assets. 

\item {\bf Crowdsourcing and research with human subjects}
    \item[] Question: For crowdsourcing experiments and research with human subjects, does the paper include the full text of instructions given to participants and screenshots, if applicable, as well as details about compensation (if any)? 
    \item[] Answer: \answerNA{} 
    \item[] Justification: The paper does not involve crowdsourcing nor research with human subjects.

\item {\bf Institutional review board (IRB) approvals or equivalent for research with human subjects}
    \item[] Question: Does the paper describe potential risks incurred by study participants, whether such risks were disclosed to the subjects, and whether Institutional Review Board (IRB) approvals (or an equivalent approval/review based on the requirements of your country or institution) were obtained?
    \item[] Answer: \answerNA{} 
    \item[] Justification: The paper does not involve crowdsourcing nor research with human subjects.

\item {\bf Declaration of LLM usage}
    \item[] Question: Does the paper describe the usage of LLMs if it is an important, original, or non-standard component of the core methods in this research? Note that if the LLM is used only for writing, editing, or formatting purposes and does \emph{not} impact the core methodology, scientific rigor, or originality of the research, declaration is not required.
    \item[] Answer: \answerNA{} 
    \item[] Justification: The core method development is totally finished by human, without any reliance on LLMs. 

\end{enumerate}

\end{document}